\documentclass[lettersize,journal]{IEEEtran}
\usepackage{amsmath,amsfonts}
\usepackage{algorithmic}
\usepackage{algorithm}
\usepackage{array}
\usepackage{textcomp}
\usepackage{stfloats}
\usepackage{url}
\usepackage{verbatim}
\usepackage{graphicx}
\usepackage[numbers]{natbib}

\usepackage{amssymb} 
\usepackage{booktabs}       
\usepackage{subcaption}
\usepackage{wrapfig}
\usepackage{xcolor}
\usepackage{balance}
\usepackage{hyperref}

\begin{document}

\title{A Bayesian Framework for Clustered Federated Learning}

\author{%
  Peng Wu, Tales Imbiriba, Pau Closas \\

}




\maketitle

\begin{abstract}
 One of the main challenges of federated learning (FL) is handling non-independent and identically distributed (non-IID) client data, which may occur in practice due to unbalanced datasets and use of different data sources across clients. Knowledge sharing and model personalization are key strategies for addressing this issue. Clustered federated learning is a class of FL methods that groups clients that observe similarly distributed data into clusters, such that every client is typically associated with one data distribution and participates in training a model for that distribution along their cluster peers. In this paper, we present a unified Bayesian framework for clustered FL which associates clients to clusters. Then we propose several practical algorithms to handle the, otherwise growing, data associations in a way that trades off performance and computational complexity. This work provides insights on client-cluster associations and enables client knowledge sharing in new ways. The proposed framework circumvents the need for unique client-cluster associations, which is seen to increase the performance of the resulting models in a variety of experiments.
\end{abstract}


\section{Introduction}
\label{intro}
\IEEEPARstart{F}{ederated} learning (FL) is a distributed machine learning approach that enables model training on decentralized data located on user devices like phones or tablets. FL allows collaborative model training without data sharing across clients, thus preserving their privacy \citep{mcmahan2017communication}. FL has been applied to computer vision \citep{10234425, liu2020fedvision}, smart cities \citep{zheng2022applications,  khan2021federated, park2022federated}, threat detection \citep{wu2023jammer}, among other pervasive applications \citep{rieke2020future}. However, FL faces significant challenges in handling non-independent and identically distributed (non-IID) data, where clients have unbalanced and statistically heterogeneous data distributions \citep{kairouz2021advances, li2020federated, li2022federated}. This violates common IID assumptions made in machine learning and leads to poor model performance.

To overcome non-IID challenges, recent works explored personalizing models to each client while still sharing knowledge between clients with similar distributions \citep{9743558, huang2021personalized, wu2021personalized}. One such approach is clustered federated learning (CFL), which groups clients by similarity and associates each client with a model, which is trained based on the data of clients on the same cluster \citep{ma2022convergence, ghosh2020efficient, long2022multi}. It usually performs better in non-IID data, with clustered clients sharing similar data distribution.
%
However, fundamental questions remain open on how to optimize client-cluster associations and inter-client knowledge sharing under non-IID data. This work aims at addressing two problems of current CFL schemes.
$1)$ clients are clustered into non-overlapping clusters, such that the participating clients tend to converge rapidly into the same cluster after several rounds of training, which may lead to inefficient utilization of local information; 
and $2)$ a lack of a unified theory to describe the information sharing among clients and their contribution to training multiple models. 


In this work, we propose a new Bayesian framework that formalizes CFL as a Bayesian data association problem \citep{bar1990tracking}, which is a process commonly used in computer vision, robotics, and data analysis that involves matching observations or measurements to the objects or sources among multiple candidates or tracks. This process is essential in multi-target tracking \citep{granstrom2017likelihood}, sensor fusion \citep{dallil2012sensor}, and situation awareness applications, where multiple objects or events are observed over time, and it is necessary to maintain consistent labeling or identification of those objects as new data arrives.
In our case, the analogy is that the global model shared by the server is treated as a mixture distribution, where each model component corresponds to an ``object'', and the local data at each client is the ``observation''.
Instead of clustering clients and associating their data to a given mode, the conceptual solution proposed here keeps track of all possible client-cluster associations through probabilistic data association. 
This results in a principled theory to model both client-cluster and client-client relations in CFL offering theoretical insights into how to optimize CFL under non-IID settings, as well as connections to existing CFL methods. 
However, the full posterior is generally intractable due to the quickly growing number of data associations over communication rounds. To address this challenge, the paper provides three practical algorithms, each leveraging different approximations related to which association hypotheses to keep at each Bayesian recursive update. A variety of experiments, including both feature and label-skew non-IID situations, show the systematic superiority of the proposed methods under the general Bayesian CFL (BCFL) framework introduced in this paper. Thus, to summarize the contribution of this work: 

1) \textbf{Novel Bayesian Framework}: This paper introduces a Bayesian framework that reinterprets Clustered Federated Learning (CFL) as a Bayesian data association problem, adapting techniques from fields like multiple target track to federated learning.

2) \textbf{Efficient Hypothesis Management}: We propose three strategies—BCFL-G, BCFL-C, and BCFL-MH—to manage the explosion of association hypotheses, balancing computational efficiency and performance.

3) \textbf{Superior Performance on Non-IID Data}: Our methods outperform existing CFL algorithms, in both feature-skew and label-skew non-IID data settings, as demonstrated by comprehensive experiments.

4) \textbf{New Research Direction}: This work reframes personalized and clustered FL as a cluster-client association problem, offering a new paradigm that could inspire further algorithmic innovations in federated learning.

\section{Related works}
\label{related_works}
\paragraph{Personalized Federated Learning.}
Many approaches try to tackle the non-IID issue in FL. Personalized FL has attracted much attention, it customizes models to each client's local data distribution. There are several ways to conduct customization. Local fine-tuning \citep{ben2010theory, wang2019federated}, meta-learning \citep{fallah2020personalized, jiang2019improving}, transfer learning \citep{li2019fedmd}, model mixture methods \citep{deng2020adaptive}, and pair-wise collaboration method \citep{huang2021personalized}. 

\paragraph{Clustered Federated Learning.}
However, these methods focus on client-level that does not consider any cluster structure, such that clients with similar backgrounds or data distributions are very likely to make similar decisions. 
%
%
%
Therefore, CFL was proposed as an alternative, which provides a middle ground by grouping similar clients that train a model on its cluster distribution \citep{mansour2020three, briggs2020federated, sattler2020byzantine, 9174245}. This balances personalization with knowledge transfer between related clients. 
Some of the existing works group the clients by model distance \citep{long2022multi, ma2022convergence} and gradient similarity \citep{duan2020fedgroup}. 
Other works utilize the training loss to assign a client to a cluster 
\citep{ghosh2020efficient, mansour2020three}. However, 
these CFL methods partition the clients into non-overlapping clusters, typically converging to the same cluster after only a few rounds of training, which results in an inefficient utilization of information for the entire training process. Other works tackle this problem by relaxing the assumption that each client can only be associated with one data distribution, called Soft Clustered FL. Some works use Fuzzy k-means \citep{bezdek2013pattern} to group clients into clusters allowing overlapping \citep{yoo2022fuzzy,stallmann2022towards,wang2023federated}, while some works assign a client to multiple clusters by comparing the inference loss either locally \citep{li2021federated} or server-side \citep{morafah2023flis}. There are also some works that assume a mixture of distributions in local clients, enabling local clients to be assigned to multiple clusters by using EM algorithm \citep{marfoq2021federated,ruan2022fedsoft}.

\paragraph{Bayesian clustered Federated Learning.}
While those works made substantial progress in different CFL directions, there is a lack of a unifying theory. This article provides a Bayesian interpretation of CFL, where client-cluster assignments are modeled using data association theory \citep{6916035,de2008new}. This principled approach enables the design of practical solutions for CFL, some of which have interpretations in connection to the existing works.
This study addresses the challenge of defining the general association between clients and clusters. While models such as IFCA \citep{ghosh2020efficient} consider non-overlapping memberships between clients and clusters, and soft CFL permits overlaps, their assignment strategies rest on heuristic choices. 
In contrast, our work can accommodate various assignment choices derived from data association theory, proposing specific, approximate assignment solutions within this theoretical framework. To enhance the generality and clarity of our approach, we adopt a Bayesian perspective to articulate the learning problems with multiple clusters and the association relationship among the clients and clusters. Please note that in Section \ref{sec:strategies}, the proposed approximate methods also rely on the assumption of non-overlapping clusters. However, we expand on this by utilizing multiple hypotheses rather than solely depending on the greedy approach, which is common in other CFL methods. 
\paragraph{CFL vs BCFL.}
Compared to CFL, Bayesian CFL enhances CFL by leveraging the benefits of Bayesian inference. By integrating prior knowledge and inferring parameter distributions, this approach effectively captures the intrinsic statistical heterogeneity of CFL data, which facilitates the quantification of uncertainties and model dynamics. Consequently, it fosters the development of more robust and interpretable federated models \citep{cao2023bayesian}. While the majority of existing Bayesian FL research has concentrated on Bayesian training employing Variational Inference \citep{corinzia2019variational,kassab2022federated}, Laplace's approximation \citep{liu2021bayesian}, or Bayesian model aggregation \citep{wu2022bayesian}. However, papers combining Bayesian FL with data association mechanisms are notably absent, especially in clustered FL. Addressing this gap is the primary contribution of our paper.

\begin{figure}[t]
    \centering
    \includegraphics[width=0.47\textwidth]{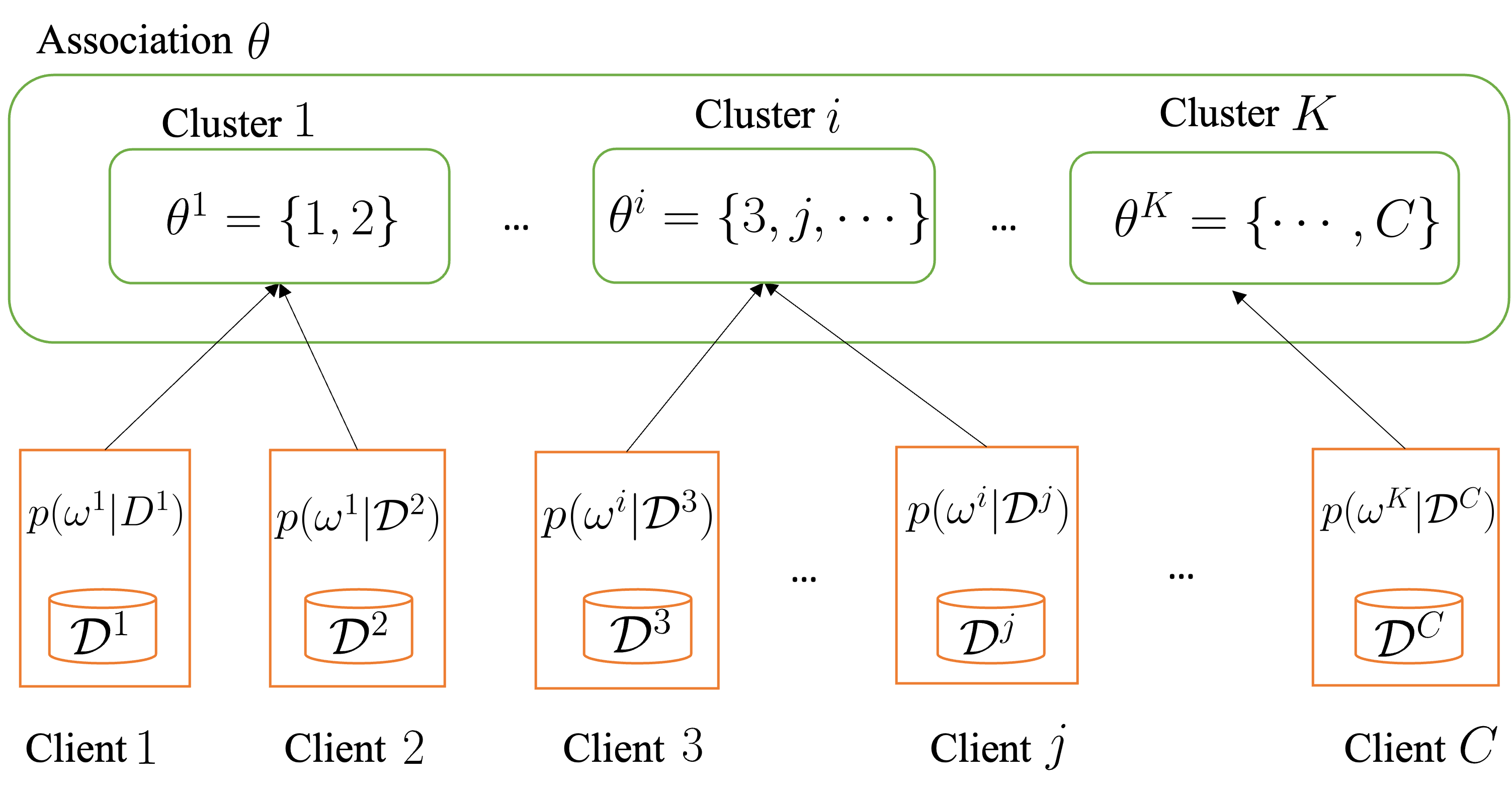}
    \includegraphics[width=0.47\textwidth]{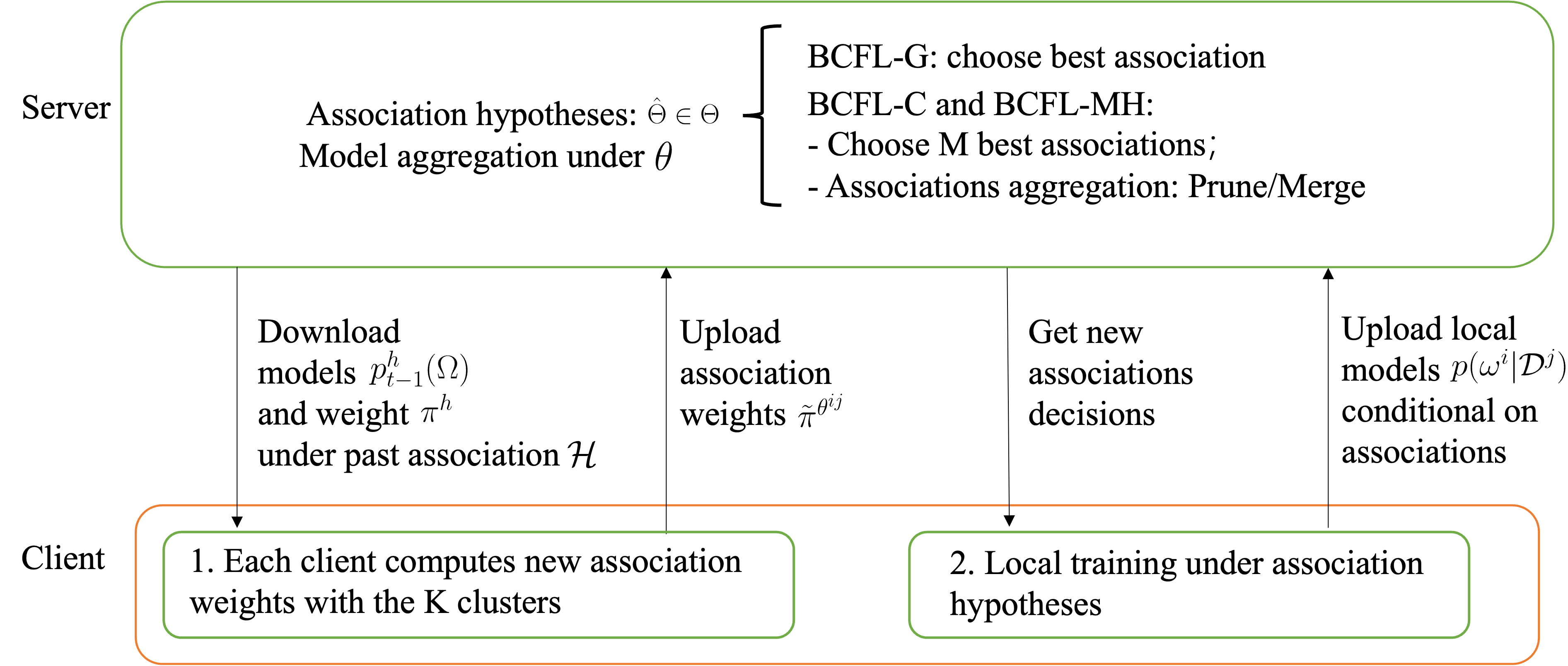}
    
    \caption{An example (top) association relation between clients and clusters; and details (bottom) of the operations and communications at the server and clients for BCFL.}
    \label{fig:framework}
\vspace{-0.3cm}
\end{figure}

\section{Bayesian solution for clustered Federated Learning}
\label{headings}
We envisage (see Figure \ref{fig:framework}) a FL system with $C$ clients, groups of which observe data drawn from similar distributions. The server aggregates local model updates in a way that generates multiple models (each corresponding to different client clustering associations) which are then shared with the clients for further local training. The proposed BCFL approach does not know the client associations beforehand, instead these are learned leveraging probabilistic data association. 
The remainder of the section presents the BCFL framework, featuring a solution that accounts for all possible associations; then we discuss a recursive update version where at each FL communication round the model updates only require local updates from the current dataset; finally, we show that such conceptual solution is generally intractable due to the growing number of associations, which motivates the approximations and practical algorithms proposed in Section \ref{sec:strategies}. For convenience, we describe the notation conventions in table \ref{tb:notation}.

\begin{table*}[htb]
\caption{Table of relevant notation conventions.}
\centering
\begin{tabular}{l|l|l}
 & Notation & Definition                                 \\ \hline
 & $K$, $i$        & Number of clustered models in server and cluster model index.      \\
 & $C$, $j$        & Number of local clients and client index.            \\
 & $\omega^i$, $\Omega$  & $i-th$ cluster model parameters and combined parameters for the $K$ clusters. \\
 & $t$ & FL communication round index \\ 
 & $\mathcal{D}^j$, $\mathcal{D}$ & $j$-th client local dataset and combined dataset for all clients. \\
 & $\theta^i$, $\theta$, $\Theta$     &  Client association indices for cluster $i$, combined association hypotheses for all \\
 & & clusters, and set of all associations.  \\
 &  $\pi^\theta$, $\tilde{\pi}^\theta$   &  Normalized and unnormalized posterior weights for association $\theta$.      \\
 &  $\pi^{j,i}$, $\tilde{\pi}^{j,i}$   &  Normalized and unnormalized local likelihood of client $j$ given cluster $i$.      \\
 &  $\pi^{j,i}_{t}$, $\tilde{\pi}^{j,i}_{t}$   &  Normalized and unnormalized local likelihood of client $j$ given cluster $i$ in communication round $t$.      \\
 & $p^\theta(\Omega)$, $p(\Omega)$     &  Posterior density given the association hypothesis $\theta$, and full posterior.      \\
 & $p^{\theta^i}(\omega^i)$, $\pi^{\theta^i}$     &  Posterior density of cluster $i$ given the association hypothesis $\theta$ and its weight.      \\
 & $\mathcal{D}^j_{1:t}$, $\mathcal{D}_{1:t}$     &  Client $j$ dataset from iterations $1$ to $t$, and combined dataset from all clients.        \\
 & $\Theta_{1:t}$, $\theta_{1:t}$     &  Set of all possible hypotheses from $1:t$ and a particular association.     \\
 & $\pi^{\theta_{1:t}}$, $p^{\theta_{1:t}}_t(\Omega)$        &  Weights and posterior given association $\theta_{1:t}$.          \\
 & $\mathcal{H}_{t-1}$, $h$         &  Set of past associations and a particular one.         \\
 &  $\pi^h$     & Accumulated weights from past $h$ associations. \\
 & $N(K,C)$ & Number of total associations given $K$ clusters and $C$ clients.               \\
 & $A$, $L$         &  Assignment and cost association matrices.         \\
 & $\theta_t^\star|h$         &  Optimal association at $t$ given past associations $h$.         \\
 & $\theta_{t,(m)}^{\star}$         &  $m$-th best  association at $t$ given past associations $h$.        \\
 & $\ell^\theta$         &  Negative posterior log-weight given $\theta$.          \\
 & $p_t(\Omega)$     &  Full posterior at $t$.     \\
 & $p_t^{\mathrm{C}}(\Omega)$     &  BCFL-C posterior approximation at $t$.     \\
 & $p_t^{\mathrm{G}}(\Omega)$     &  BCFL-G posterior approximation at $t$.     \\
 & $p_t^{\mathrm{MH}}(\Omega)$     &  BCFL-MH posterior approximation at $t$.  
 \end{tabular}
 \label{tb:notation}
\end{table*}

\subsection{Bayesian clustered Federated Learning (BCFL)}\label{sec:BCFL}
\paragraph{Problem formulation and conceptual solution.}
Following a Bayesian paradigm, our ultimate goal is to obtain the posterior distribution $p(\Omega|\mathcal{D})$ of the set of model parameters for the $K$ clusters of the server model, $\Omega\triangleq\{\omega^1, \ldots,\omega^K\}$, given the local data, $\mathcal{D} \triangleq \{\mathcal{D}^{1}, \ldots,\mathcal{D}^C\}$, of all $C$ clients. Furthermore, within the clustered FL philosophy, we aim to associate clients with each of the $K$ clusters (or models). Thus, let $\theta = \{\theta^1,\ldots,\theta^K\} \in \Theta$ be the set of $K$ associations between clients and clusters, where $\theta^i$ is a random finite set (RFS) containing the indexes of all clients associated to cluster $i$, and $\Theta$ be the set of all possible client-cluster associations. Here, $\theta^i$ is treated as a RFS since its cardinality $|\theta^i|$ and its index content are random \citep{mahler2007statistical,daley2003introduction}. With these definitions, we can expand the posterior distribution using Bayes rule and marginalization of the hypotheses as
\begin{align}\label{eq:bayesian_problem1}
    p(\Omega|\mathcal{D})
     \propto p(\mathcal{D}|\Omega) p(\Omega)
     = \sum_{\theta \in \Theta} p(\mathcal{D},\theta|\Omega) p(\Omega)  
\end{align}

\noindent where $p(\Omega)$ is the model shared by the server at a given iteration, we provide additional details regarding the recursive updates in Section \ref{sec:BCFL_rec}. 
In the paper we sometimes use $\mathcal{D}^\theta$ to compactly denote the joint variable $\{\mathcal{D},\theta\}$ which associates the data in $\mathcal{D}$ according to hypothesis $\theta$.

Besides the complexity associated with the cardinality of $\Omega$, we also aim at a posterior expression that can be evaluated in a distributed manner by fusing local posterior updates from the clients. 
Alternatively, and without further approximations, we aim to express \eqref{eq:bayesian_problem1} as a mixture distribution where each mode in the mixture represents different client-cluster association hypotheses and which can be (recursively) updated in a more manageable way. More precisely, we target a posterior characterization of the form
\begin{equation}\label{eq:general_mot}
    p(\Omega|\mathcal{D})    = \sum_{\theta \in \Theta}  \pi^{\theta}p^{\theta}(\Omega) \;, 
\end{equation}
\noindent where for a given association hypothesis $\theta$ we define 
\begin{align}
    \pi^{\theta} &= \frac{\tilde{\pi}^{\theta}}{\sum_{\theta \in \Theta} \tilde{\pi}^{\theta}} , \;\;
    \tilde{\pi}^{\theta} = \int p(\mathcal{D},\theta|\Omega) p(\Omega) d\Omega 
\label{eq:posterior_mixture_pi} \\
    p^{\theta}(\Omega) & = p(\mathcal{D},\theta|\Omega) p(\Omega)/\tilde{\pi}^\theta  \;. \label{eq:posterior_mixture_mixing}
\end{align}

\noindent Note that to guarantee that~\eqref{eq:general_mot} is a mixture of proper distributions, we use a normalization trick in \eqref{eq:posterior_mixture_pi} and \eqref{eq:posterior_mixture_mixing} where we normalized the term $p(\mathcal{D},\theta|\Omega) p(\Omega)$ with $\tilde{\pi}^\theta$.
%
Moreover, an equivalent representation of the posterior is  $p(\Omega|\mathcal{D})    = \sum_{\theta \in \Theta} p(\Omega|\mathcal{D},\theta) \mathbb{P}[\theta | \mathcal{D}]$, which provides an interpretation for the terms in \eqref{eq:general_mot} as $1)$ the weights $\pi^{\theta}$ represent the probability of a data association $\theta$ given available data, $\pi^{\theta} = \mathbb{P}[\theta | \mathcal{D}]$; and $2)$ the mixing distributions are the posterior updates given the associations $\theta$, $p^{\theta}(\Omega) = p(\Omega|\mathcal{D},\theta)$. 
We shall see that, under the assumptions that the clusters are mutually independent (i.e., $p(\Omega)=\prod_i^K p(\omega^i)$) and that the local datasets are conditionally independent ($\mathcal{D}^i \perp \mathcal{D}^{i^\prime}$, for $i\neq i^\prime$, given $\Omega$ and $\theta$), the mixing distributions in \eqref{eq:posterior_mixture_mixing} is 
\begin{align}\label{eq:posterior_mixture_mixing_fl}
    p^{\theta}(\Omega) 
        \propto 
        \prod_{i=1}^{K} \underbrace{p(\mathcal{D}^{\theta^i}|\omega^i)  p(\omega^i)}_{\propto p^{\theta^i}(\omega^i)} 
\end{align}
where $\mathcal{D}^{\theta^i}\triangleq \{\mathcal{D}^j\}_{j\in\theta^i}$, $p(\mathcal{D}^{\theta^i}|\omega^i)=p(\mathcal{D},\theta^i|\omega^i)$ is the likelihood of data associated according to $\theta^i$, and $p^{\theta^i}(\omega^i) = p(\omega^i|\mathcal{D}^{\theta^i})$ is the normalized posterior of the $i$-th cluster given its associations with the clients indexed by $\theta^i$. Similarly, the weights in \eqref{eq:posterior_mixture_pi} can be manipulated as
\begin{align}\label{eq:posterior_mixture_pi_fl}
    \pi^{\theta}
        &\propto 
        \prod_{i=1}^{K} \overbrace{\int p(\mathcal{D}^{\theta^i}|\omega^i)  p(\omega^i) d\omega^i}^{\tilde{\pi}^{\theta^i}} 
        = \prod_{i=1}^{K} \tilde{\pi}^{\theta^i} \;.
\end{align}

\noindent where $\tilde{\pi}^{\theta^i}$ is the unnormalized weight associated with $p^{\theta^i}(\omega^i)$.
It can be observed from \eqref{eq:posterior_mixture_mixing_fl} and \eqref{eq:posterior_mixture_pi_fl} that these quantities can be factorized, and thus computed, for each cluster. As a result, the target posterior \eqref{eq:general_mot} can be written as
\begin{equation}\label{eq:general_mot2}
    p(\Omega|\mathcal{D}) = \sum_{\theta \in \Theta}  \pi^{\theta}p^{\theta}(\Omega)
     =  \sum_{\theta \in \Theta} \prod_{i=1}^{K} \pi^{\theta^i}p^{\theta^i}(\omega^i) \;. 
\end{equation}
where $\pi^{\theta^i} = \tilde{\pi}^{\theta^i}/\sum_{\theta\in\Theta} \prod_{i=1}^{K} \tilde{\pi}^{\theta^i}$, and the normalization must be computed at the server. 

\paragraph{Decentralized computations.}
Finally, further manipulations to the terms in \eqref{eq:general_mot2} can be performed in order to show that both the mixing distribution and the weights, under an association hypothesis $\theta$, can be obtained from local posterior updates in a decentralized manner. First, following the methodology in~\cite{wu2022bayesian} we obtain that 
\begin{align}\label{eq:fed_posterior}
    p^{\theta^i}(\mathbf{\omega}^i) & \propto \prod_{j\in \theta^i} p(\mathcal{D}^j|\omega^i) p(\omega^i) 
     \propto \prod_{j\in \theta^i} p(\omega^i|\mathcal{D}^j) \,, 
\end{align}

\noindent where the product is over all clients associated with the $i$-th cluster under the current association hypothesis. Therefore, $p(\omega^i|\mathcal{D}^j)$ denotes the local posterior update of the parameters of the $i$-th cluster given the data from the $j$-th client, which can indeed be computed locally.
Secondly, we can similarly see that under certain approximations, the weights $\pi^{\theta}$ can be evaluated as product of locally computed weights $\pi^{\theta^i}$ for a specific data association hypothesis $\theta$. Unfortunately, integral and product in $\pi^{\theta^i}$ cannot be swapped in general. A simplification to achieve such distributed computation capability is to use the expected value of $p(\mathbf{\omega}^i)$, denoted as $\hat{\mathbf{\omega}}^i$, such that $\tilde{\pi}^{\theta^i}$ in \eqref{eq:posterior_mixture_pi_fl} are 
\begin{align}\label{eq:fed_weight}
\tilde{\pi}^{\theta^i}
    &= \int \prod_{j\in \theta^i}p(\mathcal{D}^j|\mathbf{\omega}^i)p(\mathbf{\omega}^i) d\mathbf{\omega}^i 
    \approx \prod_{j\in \theta^i} 
    \tilde{\pi}^{{j,i}}
\end{align} 

\noindent where $\tilde{\pi}^{j,i} = p(\mathcal{D}^j|\hat{\mathbf{\omega}}^i)$ are the unnormalized weights computed locally at the $j$-th client informing about the probability of associating its data to cluster $i$.
A further refinement is to employ multiple samples from $p(\mathbf{\omega}^i)$. 
For instance, one could randomly generated $N$ samples using importance or deterministic sampling. 
These set of $N$ samples and associated weights $\{\mathbf{\omega}^i_{\ell} , \alpha_\ell \}_{\ell=1}^N$ could then be used to approximate the integral in \eqref{eq:posterior_mixture_pi_fl} such that 
$\tilde{\pi}^{\theta^i}
    \approx \sum_{\ell=1}^N \alpha_\ell \prod_{j\in \theta^i} \tilde{\pi}^{j,i}_{\ell} $
with $\tilde{\pi}^{j,i}_{\ell} = p(\mathcal{D}^j|\mathbf{\omega}^i_{\ell})$.
Notice that this approach also enables local computation of weights to finally obtain \eqref{eq:posterior_mixture_pi_fl} at the server, which is in charge of computing the normalization constant $\sum_{\theta\in\Theta} \prod_{i=1}^{K} \tilde{\pi}^{\theta^i}$ requiring all local $\tilde{\pi}^{\theta^i}$.

\subsection{BCFL communication rounds as recursive updates}\label{sec:BCFL_rec}
FL typically involves multiple communication rounds, where the server shares its model parameters to the clients such that local updates can be performed. This process is performed repeatedly such that the model learning is improved over iterations, which we denote by index $t$. Similarly, BCFL can be implemented over multiple communication rounds and this section formulates the recursive update of the main equations described in Section \ref{sec:BCFL}, as well as the associated challenges.

Following a Bayesian framework, we focus in computing the posterior distribution $p_t(\Omega) \triangleq p(\Omega|\mathcal{D}_{1:t})$ of the set of parameters given all available data up to iteration $t$, that is, $\mathcal{D}_{1:t} \triangleq \{\mathcal{D}^{1}_{1:t}, \ldots,\mathcal{D}^C_{1:t}\}$ with $\mathcal{D}^j_{1:t} \triangleq \{\mathcal{D}^{j}_{1}, \ldots,\mathcal{D}^j_{t}\}$ for the $j\in\{1,\dots, C\}$ client. If a finite number of iterations $T$ are performed, then the iteration index is an integer $t\in[0, T]$ such that $t=0$ denotes the initialization step of BCFL. Notice that $T$ can be arbitrarily large, or even infinite-horizon when $T\rightarrow\infty$. This definition of the datasets $\mathcal{D}^j_{t}$ encompasses different situations. For instance, $\mathcal{D}^j_{t}$ could be randomly drawn from the same random local distribution or be the same dataset at every iteration. 

Our goal is to obtain a recursive update equation where at each time step $t$ the resulting posterior can be written as a mixture distribution in the form of~\eqref{eq:general_mot}, but summed over all possible association hypothesis at every time step. Thus, defining $\theta_t \in \Theta_t$ as the set of associations hypothesis selected at time $t$, the posterior $p_t(\Omega)$ can be written as: 
\begin{align}\label{eq:general_mot_rec}
    p_t(\Omega) = \sum_{\theta_{1:t} \in \Theta_{1:t}}  \pi^{\theta_{1:t}}p^{\theta_{1:t}}_t(\Omega) \;,
\end{align}

\noindent where $\Theta_{1:t} = \Theta_{1}\times\cdots\times\Theta_{t}$ is the set of all possible client/cluster associations until iteration $t$ such that $\sum_{\theta_{1:t} \in \Theta_{1:t}} = \sum_{\theta_{1} \in \Theta_{1}} \sum_{\theta_{2} \in \Theta_{2}} \cdots \sum_{\theta_{t} \in \Theta_{t}}$.
Since \eqref{eq:general_mot_rec} can be interpreted as $p_t(\Omega)    = \sum_{\theta_{1:t} \in \Theta_{1:t}} p(\Omega|\mathcal{D}_{1:t},\theta_{1:t}) \mathbb{P}[\theta_{1:t} | \mathcal{D}_{1:t}]$, analogously as in \eqref{eq:general_mot}, we have that: $1)$ $\pi^{\theta_{1:t}} = \mathbb{P}[\theta_{1:t} | \mathcal{D}_{1:t}]$ is the data association hypotheses posterior for the entire sequence up to iteration $t$; and 
$2)$ $p^{\theta_{1:t}}_t(\Omega) = p(\Omega|\mathcal{D}_{1:t},\theta_{1:t})$ is the model parameter posterior using data up to $t$ and association hypotheses $\theta_{1:t}$.
For a more compact notation, let us use $h \triangleq \theta_{1:t-1}$ and $\mathcal{H}_{t-1}\triangleq \Theta_{1:t-1}$ to denote a particular choice of past associations and the set of all possible past associations, respectively, such that $\Theta_{1:t} = \mathcal{H}_{t-1}\times\Theta_{t}$.

\begin{align}
    p_t(\Omega) & \propto p(\mathcal{D}_t|\Omega) p_{t-1}(\Omega) \nonumber\\
    &= \left(\sum_{\theta_{t} \in \Theta_{t}} p(\mathcal{D}_t,\theta_t|\Omega)  \right) \left(\sum_{h \in \mathcal{H}_{t-1}} \pi^{h}p^{h}_{t-1}(\Omega) \right) \nonumber\\
 & = \sum_{h \in \mathcal{H}_{t-1}} \sum_{\theta_{t} \in \Theta_{t}} \pi^{h} p(\mathcal{D}_t,\theta_t|\Omega) p^{h}_{t-1}(\Omega) \label{eq:recursive_posterior}
\end{align} 

\noindent which updates every hypothesis $h\in \mathcal{H}_{t-1}$ of the posterior at $t-1$ with every new association hypothesis $\theta_t\in\Theta_t$.
Finally, to express~\eqref{eq:recursive_posterior} as a mixture of proper distributions, one can resort to a normalization trick akin to the one used in~\eqref{eq:posterior_mixture_pi}--\eqref{eq:posterior_mixture_mixing}, reaching the mixture form in~\eqref{eq:general_mot_rec}. 
This can be achieved by considering the Bayes update $p^{\theta_{1:t}}_t(\Omega) = p(\mathcal{D}_t,\theta_t|\Omega) p^{h}_{t-1}(\Omega) / \pi^{\theta_t|h}$ under association hypotheses $\theta_{1:t}$ and corresponding unnormalized weight 
\begin{equation}\label{eq:weight_update}
    \tilde{\pi}^{\theta_{1:t}} = \pi^{h} \tilde{\pi}^{\theta_t|h}
\end{equation}
%

\noindent with $\tilde{\pi}^{\theta_t|h} = \int p(\mathcal{D}_t,\theta_t|\Omega) p^{h}_{t-1}(\Omega) d\Omega$. The normalized weights are then $\pi^{\theta_{1:t}} = \tilde{\pi}^{\theta_{1:t}} / \sum_{h \in \mathcal{H}_{t-1}} \sum_{\theta_{t} \in \Theta_{t}} \tilde{\pi}^{\theta_{1:t}}$.
We call attention for the recursive form of the weight updates in~\eqref{eq:weight_update}, where the unormalized weight $\tilde{\pi}^{\theta_{1:t}}$ is obtained by updating past hypotheses weight $\pi^h$ with the current unormalized weight hypothesis $\tilde{\pi}^{\theta_t|h}$.
Note that 
the posterior update at $t$ can be computed as detailed in Section \ref{sec:BCFL} where it is shown that this can be achieved using only local computations, which are aggregated at the server. 
{Algorithm~\ref{alg:BCFL} presents the pseudo-code for the BCFL approach, more details you can check the appendix A.}

\subsection{The explosion of association hypotheses}
\label{sec:explosion}

A major issue regarding the Bayesian framework presented above is the quick growth of association hypothesis over communication rounds $t$.  
At a given iteration, the number of possible associations for $K$ clusters and $C$ clients is given by $N(K,C) = \prod_{i=1}^K \sum_{c=0}^{C} \mathcal{C}(C,c)$, where $\mathcal{C}(C,c) = \frac{C!}{c!(C-c)!}$ represents the number of $c$ element combinations out of $C$ elements.

Furthermore, the number of hypotheses in the posterior distribution increases very rapidly due to the recursive training. 
That is, at a given iteration $t$, the number of possible clusters corresponds to the modes of the shared prior distribution, $N_t(K_{t-1},C) = |\mathcal{H}_{t-1}|= \prod_{\tau=1}^t N_\tau(K_{\tau-1},C)$, causing $N(K_t,C)$ to explode. 
Therefore, due to this \textit{curse of dimensionality}, we observe that evaluation of the exact posterior is intractable and that approximations are necessary in order to design efficient algorithms that approximate the conceptual solution provided by \eqref{eq:general_mot_rec}. The proposal of such algorithms, based on three different approximations, is the purpose of Section \ref{sec:strategies}.
In general, we aim at finding a subset of data associations $\hat{\Theta}_t \subset \Theta_t$, such that $|\hat{\Theta}_t| \ll |\Theta_t|$ while a cost minimization ensures relevant associations are kept, thus formulating this as an optimization problem. 
\section{Approximate BCFL: simplified data association strategies}\label{sec:strategies}

To address the intractability of the number of association hypotheses, this section presents three different strategies to select subsets of associations hypotheses, as well as their corresponding practical algorithms. In particular, this section discusses the metric used to quantify the cost of associating a client to a cluster, which is then used to make decisions on the desired association subsets $\hat{\Theta}_t$. 
As discussed in the literature review, Cluster Federated Learning frameworks incorporate distinct methods for assigning clients to clusters. This assignment can be seen as the one of our association here. In standard clustered FL, each client is assumed to belong to a unique, non-overlapping cluster. Cluster-client associations are determined by strategies that either rely on loss metrics or model similarity to identify the optimal alignment. However, these approaches typically restrict themselves to a single "best" association without considering how alternate associations might also benefit the overall system's performance. In contrast, soft cluster FL permits client membership across multiple, potentially overlapping clusters, yet it still defaults to a solitary association decision. Our study proposes exploring the potential advantages of allowing multiple concurrent cluster-client associations to enhance the efficacy of the federated learning process.

We examine non-overlapping structures due to the increased complexity associated with overlapping structures. To articulate this more formally:
\textbf{a key assumption 
is that at any iteration $t$ a client can only be associated with one cluster for a given hypothesis}. Notice however that since multiple association hypotheses are considered, every client has a chance to being associated with different clusters. This implies that there are no duplications in the different client partitions, thus, dramatically reducing the number of possible associations. Additionally, we assume that every selected client must be associated with a cluster such that all data is used.

\subsection{Data association as an optimization problem}\label{sec:Data association as optimization problem}

In order to simplify the overwhelmingly large number of hypotheses, this article proposes three approximate methods that select suboptimal data associations following a probabilistic approach. 
The upcoming subsections present three alternative algorithms that exploit such hypotheses reduction schemes, while in this subsection we introduce the concepts and formulate the problem as an optimization problem that can be solved efficiently.

The selection of data hypotheses is based on the maximization of their posterior probability given available data, summarized by $\tilde{\pi}^{\theta_t,h} \triangleq \pi^{h} \tilde{\pi}^{\theta_t|h}$. To avoid numerical instabilities, it is advisable to use the negative log-probabilities instead, $\ell^{\theta_t,h} \triangleq - \log ( \pi^{h} \tilde{\pi}^{\theta_t|h} )$. 
In the general case, at a given instant $t$ we aim at selecting the $M\geq 1$ hypotheses with largest (log-)probability:
\begin{equation}\label{eq:murty_assig}
    \tilde{\pi}^{\theta_{t,(1)}^{\star},h_{(1)}^{\star}} \geq \cdots \geq \tilde{\pi}^{\theta_{t,(M)}^{\star},h_{(M)}^{\star}} \geq \tilde{\pi}^{\theta,h} \;, 
\end{equation}

\noindent $\forall \theta\in\Theta_t\backslash\{\theta_{t,(m)}^{\star}\}_{m=1}^M$, $\forall h \in\mathcal{H}_{t-1}\backslash\{h_{(m)}^{\star}\}_{m=1}^M$, and where $\theta_{t,(m)}^{\star}$ and $h_{(m)}^{\star}$ denote the current and past associations leading to $m$-th highest weight, respectively.

In terms of finding the $M$ best hypotheses, we formulate this as an optimization problem such that for a given solution we aim at minimizing the negative log-weights $\ell^{\theta_t,h} = - \log ( \tilde{\pi}^{\theta_t, h} )$.
\begin{align}\label{eq:DA_optimization}
    (\theta_t^\star, h^\star) &= \mathop{\arg\min}_{(\theta_t\in\Theta_t, h\in \mathcal{H}_{t-1})} \ell^{\theta_t,h} \nonumber\\
    &= \mathop{\arg\min}_{(\theta_t\in\Theta_t, h\in \mathcal{H}_{t-1})}  \underbrace{\sum_{i=1}^K \sum_{j\in \theta^i_t} - \log ( \tilde{\pi}^{{j,i}|h} )}_{\tilde{\pi}^{\theta_t | h}} - \log \pi^h 
\end{align}

\noindent with the quantities defined in \eqref{eq:posterior_mixture_pi_fl} and \eqref{eq:fed_weight}. We observe that the optimization problem in \eqref{eq:DA_optimization} can be interpreted as an assignment problem, and thus solved using existing optimization methods. Additionally, to rank the $M$ solutions we can leverage Murty's algorithm \citep{miller1997optimizing} in conjunction to the assignment solution.

To connect \eqref{eq:DA_optimization} to the assignment problem formulation \citep{alfaro2022assignment} adopted in this paper, we introduce its basic notation. 
Let $L\in\mathbb{R}^{C \times K}$ be a cost matrix whose entries $L^{j,i}$ represent the cost of assigning the $j$-th client to the $i$-th cluster, and $A$ be a binary assignment matrix with entries $A^{j,i}\in\{0,1\}$. The total cost of a given association $A$ can be written as $\mathrm{Tr} (A^\top L)$. The assumption that a client's data can only be associated with one cluster can be imposed with the constraint $\sum_{i=1}^K A^{j, i}=1$, $ \forall j$ over the association matrix $A$. 
%
In general, we would like to select associations with the smallest cost. Obtaining the best association according to $\mathrm{Tr} (A^\top L) = \sum_{i=1}^{K}\sum_{j=1}^{C} A^{i,j} L^{j,i}$ can be formalized as a constrained optimization problem of the form
\begin{align}\label{eq:opt_assig}
    A^\star &= \mathop{\arg\min}_{A} \sum_{i=1}^{K}\sum_{j=1}^{C} A^{i,j} L^{j,i},  
    \nonumber  \\ 
    &\text{s.t.}\quad 
    A^{i,j}\in \{0,1\}, \text{ and } \sum_{i=1}^K A^{j,i}=1, \, \forall j \;, 
\end{align}
\noindent for which efficient algorithms exist such as the Hungarian or the Auction algorithms \citep{fredman1987fibonacci}, as well as the Jonker-Volgenant-Castanon algorithm \citep{jonker1988shortest,drummond1990comparison}. Remarkably, a naive solution to the optimization has factorial complexity, while those algorithms can solve the assignment problem in polynomial time.
The optimization of \eqref{eq:opt_assig} results in the optimal association matrix $A^\star$, which is equivalent to finding the optimal association variable $\tilde{\pi}^{\theta_t^\star,h^\star}$ where the elements of the cost matrix $L^{i,j}=- \log ( \tilde{\pi}^{{j,i}|h} ) - \log \pi^h$ are such that the assignments $A$ correspond to a unique $\theta_t\in\Theta_t$. As a consequence, we can use the aforementioned assigment algorithms to find sets of client-cluster associations, which is leveraged in the practical BCFL methods proposed next. We provide a simple example in appendix figure \ref{fig:example of assoication}.

In the next subsections we will discuss three simplification approaches to reduce the total number of associations carried from one iteration to the next. When only one hypothesis is maintained, that is, $|\mathcal{H}_{t-1}|=1$, then $\tilde{\pi}^{\theta,h}= \tilde{\pi}^{\theta|h}$ since $\pi^{h}=1$. In that scenario the optimation problem in~\eqref{eq:DA_optimization} simplifies since the $\log\pi^h=0$. Equivalently, $L^{i,j}$ becomes $L^{i,j} = - \log (\tilde{\pi}^{{j,i}|h})$ in~\eqref{eq:opt_assig}.  

\subsection{Approximate BCFL algorithms}
\textbf{Greedy Association: BCFL-G.}
The most simplistic association is to follow a greedy approach where at each iteration $t$ only the best association is kept, leading to $|\mathcal{H}_{t-1}|=1$.
%
Denoting as $\theta^\star$ the association leading to the optimal assignment, $M=1$, and denoting the sequence of optimal associations by $\theta^{\star}_{1:t}\triangleq \{\theta^\star_1, \theta^\star_2|\theta^\star_1,\ldots, \theta^\star_t|\theta^\star_{1:t-1}\}$, the posterior in~\eqref{eq:general_mot_rec} can be approximated as
\begin{align} \label{eq:greedy}
    p_{t}^{\textrm{G}}(\Omega) = p^{\theta^{\star}_{1:t}}_{t}(\Omega) 
    = \prod_{i=1}^K p^{\theta^{\star,i}_{1:t}}_{t}(\omega^i)
\end{align}

\begin{algorithm}[H]
    \centering
    \caption{Conceptual BCFL algorithm}
   \label{alg:BCFL}
\begin{algorithmic}
   \STATE {\bfseries Input:} $K$
   \STATE {\bfseries Initialization:} $\mathcal{H}_0$, $\{\{\pi^{\theta^i}_0, p^{\theta^i}_{0}(\omega^i)\}_{i=1}^K\}_{\theta \in \mathcal{H}_0}$; 
   \FOR{$t=1,\cdots,T$}
   \STATE \textbf{Server do} 
   \STATE \qquad Broadcast $\{\{p^{\theta^i|h}_{t-1}(\omega^i)\}_{i=1}^K\}_{\theta \in \mathcal{H}_{t-1}}$ to participating clients 
   \STATE \textbf{end Server} 
    \FOR{Client $j\in\{1,\dots,C\}$ in parallel}
        \FOR{every hypothesis $h \in \mathcal{H}_{t-1}$ in parallel}
            \STATE Compute association weight
            $\{\tilde{\pi}^{j,i|h}_{t}\}_{i=1}^K$ as in \eqref{eq:fed_weight}
            \STATE Compute posterior update $\{p_{t}^{\theta^i|h}(\omega^{i}|\mathcal{D}^j)\}_{i=1}^K$ for each cluster $i$
            \STATE Transmit $\{\tilde{\pi}^{j,i|h}_{t}, p_{t}^{\theta^i|h}(\omega^{i}|\mathcal{D}^j)\}_{i=1}^K$ to the server
        \ENDFOR
    \ENDFOR
    \STATE \textbf{Server do}
    \STATE \qquad Update hypothesis set $\mathcal{H}_{t}=\mathcal{H}_{t-1} \times \Theta_{t}$ 
    \STATE \qquad  Compute $\{p^{\theta^i|h}_{t}(\omega^i)\}_{i=1}^K$ as in \eqref{eq:fed_posterior} 
    \STATE \qquad Compute posterior $p^{\theta_{1:t}}_t(\Omega) \propto \prod_{i=1}^{K} p^{\theta^i|h}_t(\omega^i)$, for all $\theta_{t} \in \Theta_{t}$ and $h \in \mathcal{H}_{t-1}$
    \STATE  \qquad  Compute $\{\pi^{\theta^i|h}_{t}\}_{i=1}^K$ as in equation \ref{eq:fed_weight} and obtain ${\pi}^{\theta_{1:t}} \propto \pi^{h} \tilde{\pi}^{\theta_t|h}$
    \STATE \qquad Compute full posterior $p_t(\Omega)$ as in equation \ref{eq:recursive_posterior}
    \STATE \textbf{end Server} 
   \ENDFOR
    \STATE {\bfseries return:} $p_t(\Omega)$
\end{algorithmic}
\end{algorithm}

\begin{algorithm}[H]
    \centering
    \caption{BCFL-G algorithm }
   \label{alg:BCFL-G}
\begin{algorithmic}
   \STATE {\bfseries Input:} $K$
   \STATE {\bfseries Initialization:} $\{p^{i}_{0}(\omega^i)\}_{i=1}^K$; 
   \FOR{$t=1,\cdots,T$}
       \STATE \textbf{Server do} 
       \STATE \quad broadcast $\{p^{i}_{t-1}(\omega^i)\}_{i=1}^K\}$ to selected clients
       \STATE \textbf{end Server} 
        \FOR{Client $j\in\{1,\dots,C\}$ in parallel}
            \STATE Compute association weight $p_{t}(\mathcal{D}^j|\omega^i_1)$ 
            \STATE Compare weights across clusters and get the best assignment 
            \STATE Compute the local posterior $p^*_{t}(\omega^i|\mathcal{D}^j)$ and transmits it to the server
        \ENDFOR
        \STATE \textbf{Server do} 
        \STATE \quad Compute the cluster posteriors $\{ p^{\theta^{\star,i}_{1:t}}_{t}(\omega^i)\}_{i=1}^K$ 
        \STATE \quad Compute the posterior $p_{t}^{\textrm{G}}(\Omega)$ as in equation \ref{eq:greedy}
        \STATE \textbf{end Server}
   \ENDFOR
    \STATE {\bfseries Return:} $p_{t}^{\textrm{G}}(\Omega)$ 
\end{algorithmic}
\end{algorithm}

where we have $\pi^{\theta_{1:t}} = \pi^{\theta^\star_{1:t}} = 1$, since only the best association is kept at each time $t$. The posterior in~\eqref{eq:greedy} can be recursively obtained from the posterior of the previous time step, that is, $p^{\theta^{\star,i}_{1:t}}_{t}(\omega^i) \propto p(\mathcal{D}_{t}^{\theta^\star_t,i}|\omega^i_t)p^{\theta^{\star,i}_{1:t-1}}_{t-1}(\omega^i)$ as discussed in Section \ref{sec:BCFL_rec}. Note that unnormalized local updates of the posterior can be computed locally at the clients while aggregation and normalization need to be performed in the server. {Algorithm~\ref{alg:BCFL-G} presents the pseudo-code for BCFL-G.}

The greedy approach has several benefits, mostly related to its reduced complexity which makes it computationally cheap and relatively simple to implement. The downside is that there is no guarantee that the selected trajectory of hypotheses -- which are the best for a given iteration conditional on past associations -- is optimal in a broader sense of $\tilde{\pi}^{\theta_{1:t}^\star} \geq \tilde{\pi}^{\theta_{1:t}} \;, \forall \theta_{1:t}\in\Theta_{1:t}$. Therefore, this points out that keeping a specific association only might not be sufficient to represent the uncertainty of the random variable $\theta_{1:t}$, which motivates the next two practical strategies. 

The resulting BCFL-G, for \textit{greedy}, association strategy is somewhat correlated with the method proposed in \cite{ghosh2020efficient}, despite their deterministic perspective, in which the best clusters are selected at every time step by maximizing the data-fit at each client. However, regardless the existing similarities, we highlight that the Bayesian framework proposed in this paper is more general, which enables many other association strategies as discussed next. 

\textbf{Consensus Association: BCFL-C.}
Consensus strategies for propagating multiple hypothesis distributions have been proposed in the multi-target tracking scenario \citep{rezatofighi2015joint}. 
Analogous to the joint probabilistic data association (JPDA) tracker, we propose an association strategy that is composed of two steps. First, the best $M>1$ associations are found and the equivalent posteriors are updated. Then, these posteriors are merged leading to a single hypothesis. This implies that at every iteration the number of past hypotheses is $|\mathcal{H}_{t-1}|=1$, simplifying the optimization problem in Section~\ref{sec:Data association as optimization problem}.

We refer to this the \textit{consensus} approach, or BCFL-C for short. This additional aggregation of hypotheses improves the uncertainty characterization of $\theta_{1:t}$, which in turn results in better performance results as is discussed in the results section. Noticeably, the computational complexity of BCFL-C is comparable to that of BCFL-G.
Following similar definitions as in BCFL-G, we denote the resulting posterior approximation as
\begin{align} \label{eq:jpda_posterior}
    p_{t}^{\mathrm{C}}(\Omega) 
    = \prod_{i=1}^{K} \!\mathsf{MERGE}\!\left(\sum_{m=1}^{M} {\pi}^{\theta_{t,(m)}^{\star,i}|h} p_{t|t-1}^{\theta_{t,(m)}^{\star,i}|h}(\omega^i)\right) 
\end{align} 

\noindent where $p_{t|t-1}^{\theta_{t,(m)}^{\star,i}|h}(\omega_t^i)$ is the posterior distribution of cluster $i$ given the $m$-th best instantaneous hypothesis $\theta_{t,(m)}^{\star,i}$ and past associations $h$. The $\mathsf{MERGE}(\cdot)$ operator fuses the $M$ hypotheses into a single density,implemented by moment matching or other techniques \citep{bishop2006pattern}. {For Gaussian densities, this can be easily obtained \citep{li2019second,luengo2018efficient}, see Appendix \ref{ap:Gaussian_fusion}. Algorithm
~\ref{alg:BCFL-approx}
shows BCFL-C pseudo-code.}

\textbf{Multiple Association Hypothesis: BCFL-MH.}
A relaxation of the approximations performed by BCFL-G and BCFL-C, where a single hypothesis is propagated over iterations, is to keep track of several trajectories of possible association hypotheses, similarly proposed in the multiple hypotheses tracking (MHT) filtering target tracking \citep{kim2015multiple}. The general posterior in \eqref{eq:general_mot_rec} then results in the \textit{multi-hypothesis} approach BCFL-MH:
\begin{equation}\label{eq:mht}
    p_t^{\mathrm{MH}}(\Omega) = \sum_{\theta_{1:t} \in \hat{\Theta}_{1:t}}  \pi^{\theta_{1:t}}p^{\theta_{1:t}}_t(\Omega) \;,
\end{equation} 
\begin{algorithm}[H]
    \centering
    \caption{Approximate BCFL-C and BCFL-MH algorithms}
   \label{alg:BCFL-approx}
\begin{algorithmic}
   \STATE {\bfseries Input:} $K$, $M_{\text{max}}$
   \STATE {\bfseries Initialization:} $\mathcal{H}_0$, $\{\{\pi^{\theta^i}_0, p^{\theta^i}_{0}(\omega^i)\}_{i=1}^K\}_{\theta \in \mathcal{H}_0}$; 
   \FOR{$t=1,\cdots,T$}
       \STATE \textbf{Server do} 
       \STATE \qquad Broadcast $\{\{p^{\theta^i|h}_{t}(\omega^i)\}_{i=1}^K\}_{\theta \in \mathcal{H}_{t-1}}$ to participating clients
        \STATE \textbf{end Server} 
      \FOR{Client $j\in\{1,\dots,C\}$ in parallel}
        \FOR{every hypothesis $h\in\mathcal{H}_{t-1}$ in parallel}
            \STATE Compute association weight
            $\{\tilde{\pi}^{j,i|h}_{t}\}_{i=1}^K$ as in \eqref{eq:fed_weight}
            \STATE Transmit weights to the server
        \ENDFOR
      \ENDFOR
      \STATE \textbf{Server do}
      \STATE \qquad Construct cost matrix $L$ as in section \ref{sec:Data association as optimization problem} 
      \STATE \qquad Compose $\Theta_{t}$ by keeping the $M_{\text{max}}$ best associations 
      \STATE \qquad $\mathcal{H}_{t} = \Theta_{t}\times \mathcal{H}_{t-1}$, 
        \STATE \textbf{end Server} 
      \FOR{$\theta\in\mathcal{H}_{t}$ in parallel}
      \FOR{$i$ in $K$}
        \STATE \textbf{Server do} 
        \STATE \quad Broadcast association decision $\theta^i$ to associated clients
        \STATE \textbf{end Server} 
        \FOR{Client $j\in\theta^{i}$ in parallel}
          \STATE Compute local posterior $p_{t}^{\theta^i|h}(\omega^{i}|\mathcal{D}^j)$ and transmit to the server
        \ENDFOR
        \STATE \textbf{Server do}
        \STATE \quad Compute $p^{\theta^{i}|h}_{t}(\omega^i)$, $\pi^{\theta^{i}|h}_{t}$ using equations \ref{eq:fed_posterior} and \ref{eq:fed_weight}
        \STATE \textbf{end Server}         
        \ENDFOR
      \ENDFOR
      \STATE \textbf{Server do} 
      \STATE \quad\textbf{if} BCFL-C \textbf{then}
      \STATE \qquad Merging $M_{\text{max}}$ best hypothesis into one, $|\mathcal{H}_{t}|=1$
      \STATE \quad \textbf{end if}
      \STATE \quad\textbf{if} BCFL-MH \textbf{then}
      \STATE \qquad Pruning $M_{\text{max}}$ best hypotheses with largest posterior weights,  $|\mathcal{H}_{t}|=M_{\text{max}}$
      \STATE \quad \textbf{end if}
      
     \STATE \textbf{end Server}
    \ENDFOR
     \STATE {\bfseries Return:} $p_{t}^{\mathrm{C}}(\Omega) $ or $p_{t}^{\mathrm{MH}}(\Omega) $
\end{algorithmic}
\end{algorithm}
\noindent which in essence implies finding the subset $\hat{\Theta}_{1:t} \subset \Theta_{1:t}$ of $M_\text{max}=|\hat{\Theta}_{1:t}|$ highly-promising hypotheses that the method would update. The identification of this subset and its recursive update can be performed by pruning associations with weights smaller than a predefined threshold, then use Murty's algorithm or similar to rank the best $M_\text{max}$ associations. BCFL-MH is arguably more complex to implement due to the need for keeping track of multiple association hypotheses, more discussion of complexity analysis \cite{talbi2009metaheuristics} can find in appendix A. However, we will see that its performance is typically superior since for large $M_\text{max}$ values the uncertainty of associations hypotheses can be accurately characterized. 
The BCFL-MH distribution in \eqref{eq:mht} is then parameterized by weights and densities for each of the $M_\text{max}$ trajectories selected at round $t$, $\{ \pi^{\theta_{1:t}} , \{p^{\theta_{1:t}^i}_t(\omega^i)\}_{i=1}^K\}_{\theta_{1:t} \in \hat{\Theta}_{1:t}}$. For a given hypothesis, the weight and densities are computed as described in
Section \ref{headings} and Algorithm~\ref{alg:BCFL-approx}.

\section{Experiments}\label{sec:results}

To validate the proposed BCFL methods under both feature- and label-skew situations, we generate four non-IID scenarios using four well-known datasets: Digits-Five \citep{peng2019moment}, AmazonReview \citep{blitzer2007biographies}, Fashion-MNIST \citep{xiao2017fashion} and CIFAR-10 \citep{krizhevsky2009learning} (see Appendix C
). Digits-Five and AmazonReview contain data coming from different categories, making them suitable to test feature-skewed situations.

To generate the feature-skewed scenario we split data with different characteristics among multiple clients. We create two scenarios using Digits-Five and AmazonReview since data from these datasets are already categorized into different groups. 
For Digits-Five, we split the data among $C=10$ clients, $2$ per sub-dataset, leading to $5$ disjoint client groups. 
For AmazonReview, we split the data among $C=8$ clients, $2$ per merchandise category.  
%
As for label-skewed data, we generate two scenarios using Fashion-MNIST and CIFAR-10, generating label-skewed groups using a two-stage Dirichlet-based sampling approach~\citep{ma2022convergence,li2021federated}, process that is controlled by the concentration parameter $\alpha$ which is set to $0.1$ in our experiments.

\textbf{Baseline models and system settings.}
We selected several existing methods for benchmarking purposes: the well-known FedAvg \citep{mcmahan2017communication}, a single-model-based FL strategy; FeSEM and WeCFL, which are state-of-the-art clustered FL methods \citep{long2022multi, ma2022convergence}; and FedAMP, a popular personalized FL algorithm. We also included Per-FedAvg \citep{fallah2020personalized}, adapted to a Bayesian perspective (referred to as BP-FedAvg) to maintain a consistent training structure with our algorithms. Additionally, we compared with FedSoft \citep{ruan2022fedsoft}, a well-known Soft Clustered FL method.
In this work, we consider the training of a neural network (NN) model using FL, in which case the local posteriors in \eqref{eq:fed_posterior} are obtained using Laplace approximation as in \cite{liu2021bayesian} and the weights in \eqref{eq:fed_weight} are related to the corresponding training loss. 
System setting details can be found in Appendix \ref{app:experiments}. 
In the experiments, we also evaluate the impact of model warm-up, whose purpose is to improve the initialization of local models, potentially improving the overall FL solution as is discussed in Appendix \ref{ap:moreresults} along with extra experiments.

\textbf{Evaluation metrics:} We evaluate the performance using both micro accuracy (acc \%) and macro F1-score (F1) on the client-wise test datasets due to high non-IID degrees. Micro average is performed for accuracy to balance the different number of samples in the clients. 

\begin{table*}[htb]
\caption{Performance comparison on cluster-wise non-IID. For Digits-Five and AmazonReview feature-skewed data scenarios, Feature $(K, C/K)$ indicates $K$ data groups with $C/K$ clients per group. For Fashion-MNIST and CIFAR-10 label-skewed data scenarios, Label $(K, C/K, \alpha)$ indicates 
also the value of the Dirichlet concentration parameter $\alpha$.}

\centering
\begin{tabular}{l|cc|cc|cc|cc}
\toprule
\multicolumn{1}{c|}{Datasets} &
  \multicolumn{2}{c}{Digits-Five} &
  \multicolumn{2}{c}{AmazonReview} &
  \multicolumn{2}{c}{Fashion-MNIST} &
  \multicolumn{2}{c}{CIFAR-10} \\ \midrule
\multicolumn{1}{l|}{non-IID setting} &
  \multicolumn{2}{c}{Feature $(5, 2)$} &
  \multicolumn{2}{c}{Feature $(4, 2)$} &
  \multicolumn{2}{c}{Label $(4,10, 0.1)$} &
  \multicolumn{2}{c}{Label $(4,10, 0.1)$} \\ \midrule
Methods &
  \multicolumn{1}{c}{Acc} &
  \multicolumn{1}{c}{F1} &
  \multicolumn{1}{c}{Acc} &
  \multicolumn{1}{c}{F1} &
  \multicolumn{1}{c}{Acc} &
  \multicolumn{1}{c}{F1} &
  \multicolumn{1}{c}{Acc} &
  \multicolumn{1}{c}{F1} \\ \bottomrule\toprule
FedAvg &
  $93.18$ &
  $92.95$ &
  $87.71$ &
  $87.70$ &
  $85.00$ &
  $53.14$ &
  $39.89$ &
  $21.19$ \\ \midrule
BP-FedAvg &					
  $90.33$ &
  $90.01$ &
  $86.50$ &
  $86.44$ &
  $94.93$ &
  $85.14$ &
  $66.59$ &
  $43.23$ \\ \midrule
FedAMP &
  $92.71$ &
  $92.43$ &
  $87.90$ &
  $87.82$ &
  $95.70$ &
  $76.42$ &
  $67.22$ &
  $48.15$ \\ \midrule
FeSEM &
  $93.98$ &
  $93.80$ &
  $85.30$ &
  $85.27$ &
  $96.82$ &
  $92.77$ &
  $72.88$ &
  $51.02$ \\ \midrule
WeCFL &
  $93.81$ &
  $93.58$ &
  $85.57$ &
  $85.51$ &
  $96.74$ &
  $92.0$ &
  $72.91$ &
  $51.69$ \\ \midrule
FedSoft &
  $91.65$ &					
  $91.38$ &
  $85.85$ &
  $85.78$ &
  $93.90$ &
  $84.77$ &
  $68.01$ &
  $48.27$ \\ \midrule
BCFL-G &
  $94.35$ &
  $94.16$ &
  $87.53$ &
  $87.5$ &
  $96.81$ &
  $93.51$ &
  $64.73$ &
  $44.22$ \\ 
BCFL-C-3 &
  $94.04$ &
  $93.84$ &
  $87.95$ &
  $87.93$ &
  $96.84$ &
  $90.16$ &
  $72.12$ &
  $50.97$ \\ 
BCFL-C-6 &
  $94.14$ &
  $93.93$ &
  $88.11$ &
  $88.09$ &
  $96.58$ &
  $86.55$ &
  $68.73$ &
  $47.40$ \\ 
BCFL-MH-3 &
  $95.39$ &
  $95.22$ &
  $\textbf{88.74}$ &
  $\textbf{88.74}$ &
  $97.13$ &
  $\textbf{93.70}$ &
  $74.35$ &
  $\textbf{56.24}$ \\ 
BCFL-MH-6 &
  $\textbf{96.02}$ &
  $\textbf{95.88}$ &
  $88.16$ &
  $88.15$ &
  $\textbf{97.27}$ &
  $92.56$ &
  $\textbf{75.26}$ &
   $53.45$ \\ 
  \bottomrule
\end{tabular}
\label{tb:table1}
\end{table*}

\begin{figure}
\centering
\includegraphics[width=0.24\textwidth]{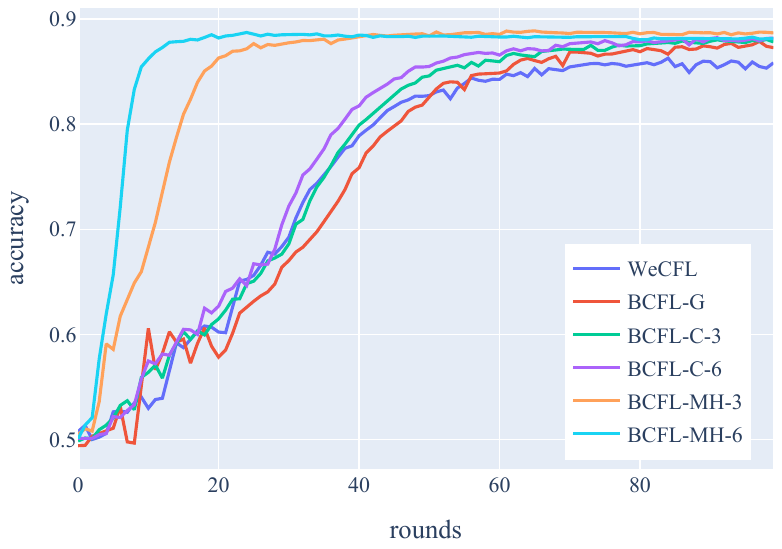}\hfill
\includegraphics[width=0.24\textwidth]{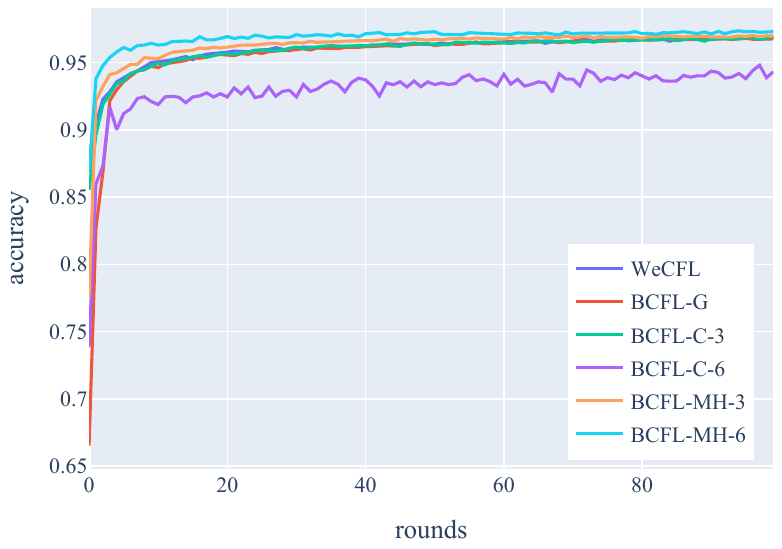}
\caption{Accuracies for AmazonReview (left panel) and Fashion-MNIST (right panel) datasets.}
\label{fig:accuracy_rounds}
\end{figure}
%
\textbf{Comparison study.}
Table \ref{tb:table1} provides insights into the performance of different methods on various datasets. 
We refer to the BCFL methods by their acronyms.
For BCFL-C and BCFL-MH, a number is also included to denote the number of associations considered, $M$ and $M_\text{max}$, respectively.
We can notice the superior performance of the proposed BCFL variants. In fact, all the best results were obtained by the BCFL-MH class of methods in all datasets. We highlight, however, that the BCFL-MH variants are inherently more costly since they explore multiple association hypotheses throughout iterations. The less expensive BCFL-G and BCFL-C present results that are close to the results obtained BCFL-MH in most datasets and slightly superior to results obtained with other algorithms
FedAvg, which serves as a baseline, is surpassed by WeCFL, FedSEM in terms of performance, except for AmazonReview dataset. 
This suggests that AmazonReview dataset may not exhibit strong non-IID characteristics, given that a centralized model like FedAvg can outperform other methods, including BCFL-G.


\textbf{Convergence analysis.}
Figure \ref{fig:accuracy_rounds} shows the convergence curves of several clustered FL methods including WeCFL, and multiple BCFL variants (G, C-3, C-6, MH-3, and MH-6) for the AmazonReview and Fashion-MNIST datasets. 
It indicates that BCFL-MH-6 exhibits the fastest convergence rate among all methods. Indeed, according to our practical experience, BCFL-MH converges faster than the other methods in general.


\textbf{Clustering analysis.}
To better understand the dynamics of information-sharing among clients, we visualize the clients' cluster graph across all rounds. Figure~\ref{fig:cluster} focuses on the Digits-Five dataset, similar analysis for the other experiments can be found in Appendix~\ref{app:fexps}. In the figure, the thickness of the edges between clients represents the degree of information sharing between them during the training process (i.e. their accumulated probability of associating to the same cluster). 

\begin{figure}[htb]
    \centering
    \begin{subfigure}[b]{0.07\textwidth}
        \includegraphics[width=\textwidth]{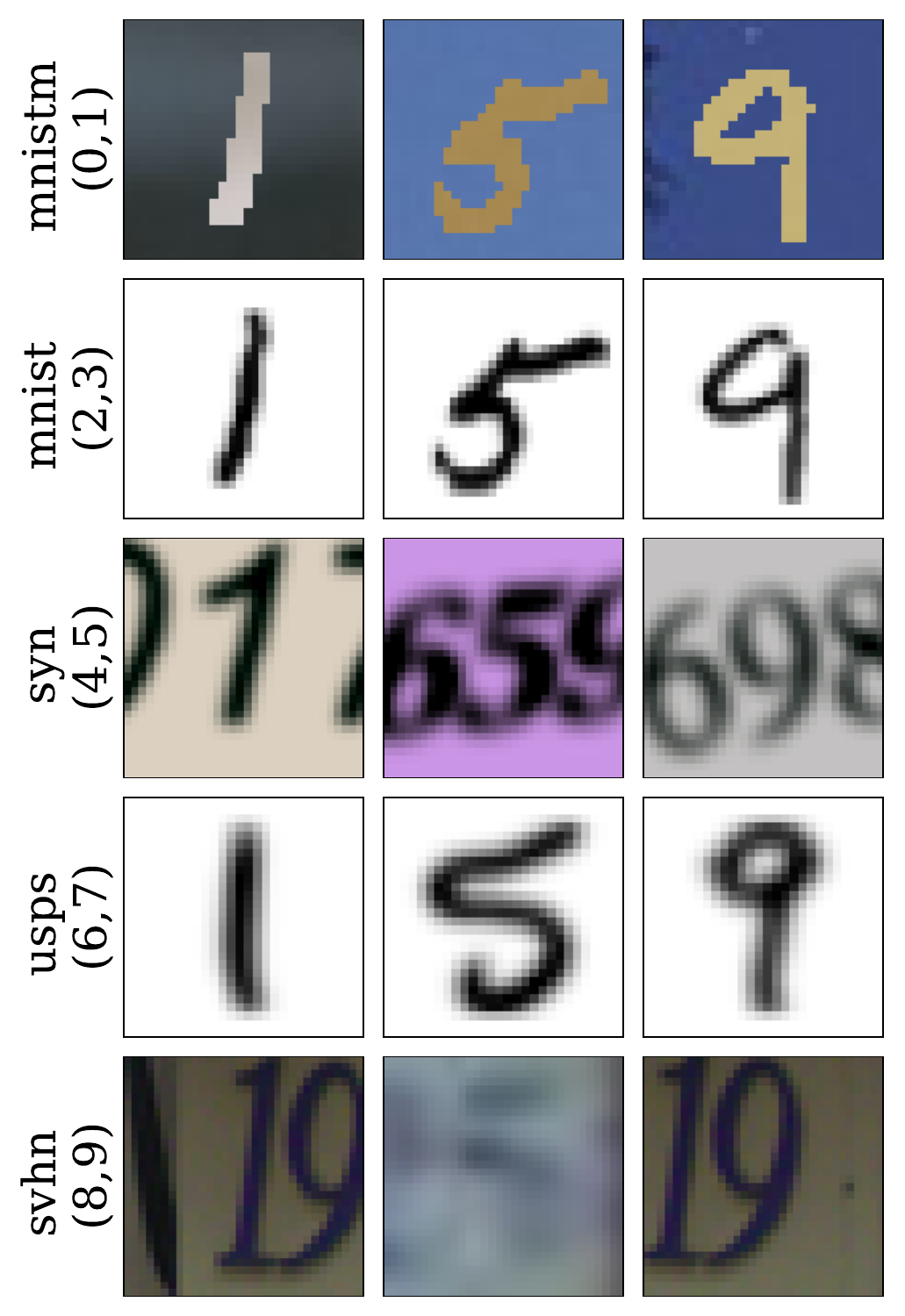}
        \caption{DF}
    \end{subfigure}
    \hfill
    \begin{subfigure}[b]{0.13\textwidth}
        \includegraphics[width=\textwidth]{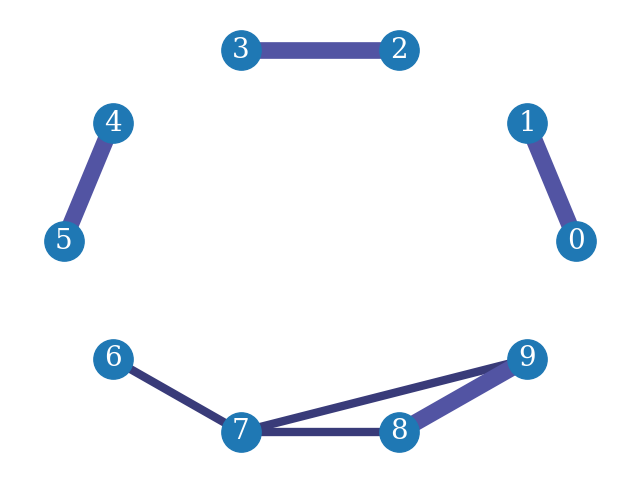}
        \caption{WeCFL}
    \end{subfigure}
    \hfill
    \begin{subfigure}[b]{0.13\textwidth}
        \includegraphics[width=\textwidth]{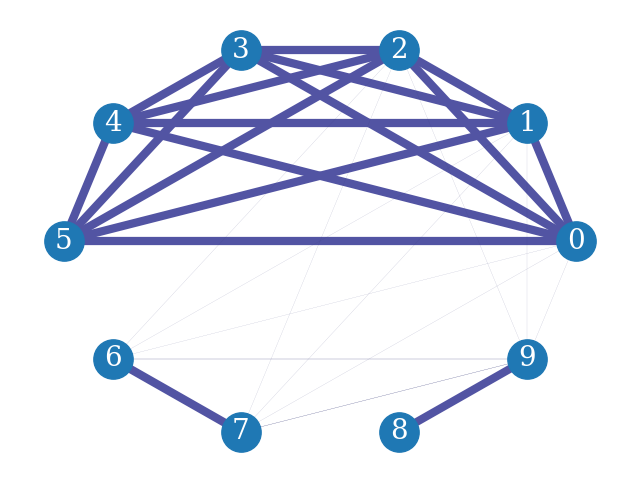}
        \caption{BCFL-G}
    \end{subfigure}
    \hfill
    \begin{subfigure}[b]{0.13\textwidth}
        \includegraphics[width=\textwidth]{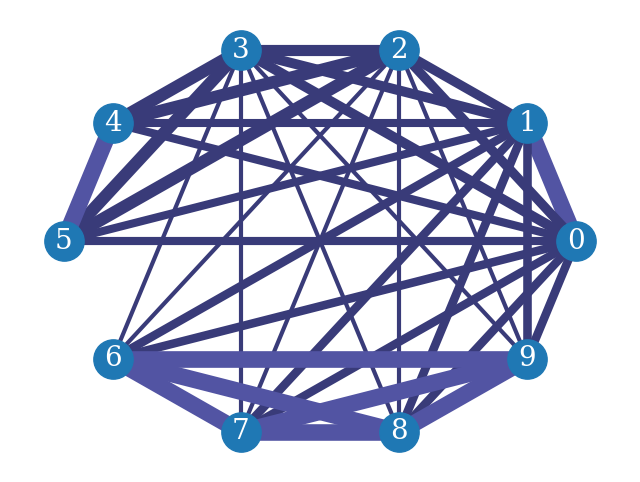}
        \caption{BCFL-MH-3}
    \end{subfigure}
    \caption{Client clustering during training for Digits-Five and selected CFL methods. $K=5$ clusters are pairwise split into $C=10$ clients, noted on the left labeling in (a).}
    \label{fig:cluster}
\end{figure}

The graph shows that while WeCFL indeed converges to groups, BCFL-G exhibits more client connections and potential for information exchange. With even more client connections, BCFL-MH formed connections among almost all clients. Nevertheless, we can notice that stronger connections were formed between client groups $\{0,1\}$, $\{2,3\}$, $\{4,5\}$ and $\{6,7,8,9\}$, correctly clustering clients observing similar features. There is a considerable connection between groups $\{2,3\}$ and $\{4,5\}$ as well, which could be considered a cluster itself.

\section{Conclusion}\label{sec:conclusion}
This work presents a unifying framework for clustered FL that employs probabilistic data association to infer the relation between clients and clusters. The framework shows the conceptual solution to the problem and highlights the need for approximations leading to feasible implementations. It paves the way for new research solutions to address the long-standing challenge of handling non-IID data in FL systems. In particular, three different approximations to reduce the number of association hypotheses are proposed, with different complexity requirements, exhibiting competitive results superior to state-of-the-art methods. 
Additionally, the probabilistic framework provides uncertainty measures that enable cross-client/cluster pollination, enhancing performance.
Future work includes extending BCFL to estimate the number of clusters, which has the added challenge that the parameter space has an unknown cardinality.

\bibliography{iclr2024_conference}
\bibliographystyle{IEEEtran}

\newpage

\section{Biography Section}
 


\begin{IEEEbiography}[{\includegraphics[width=1in,height=1.25in,clip,keepaspectratio]{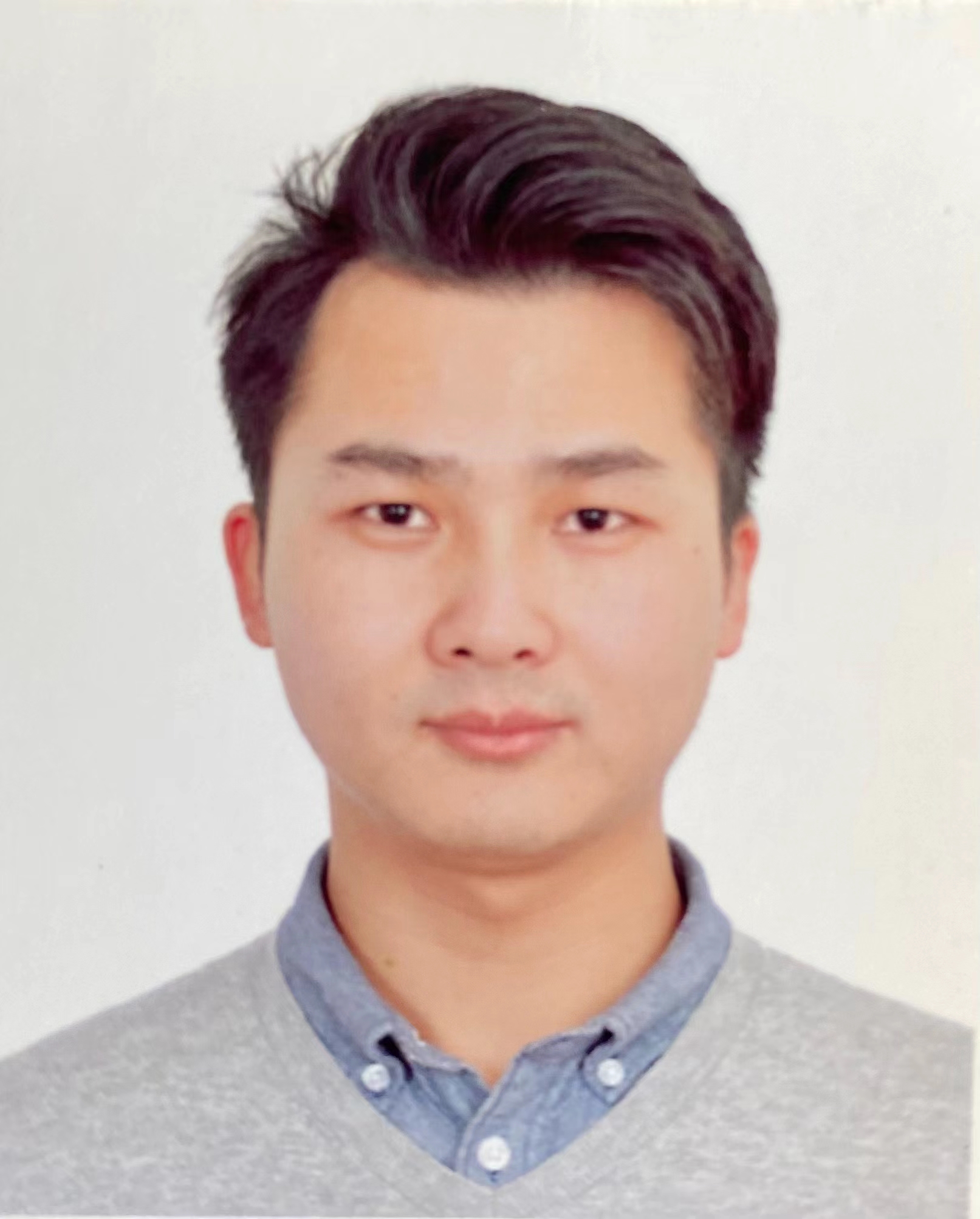}}]{Peng Wu}{\space}received his BS degree in Physics from Tianjin University of Technology, China in 2012 and MS degree in Electrical Engineering from Northeastern University, Boston, MA, in 2018. He graduated with a PhD from the Department of Electrical and Computer Engineering at Northeastern University in 2024. His research interests include distributed data fusion and machine learning with applications to indoor positioning and tracking.
\end{IEEEbiography}

\begin{IEEEbiography}[{\includegraphics[width=1in,height=1.25in,clip,keepaspectratio]{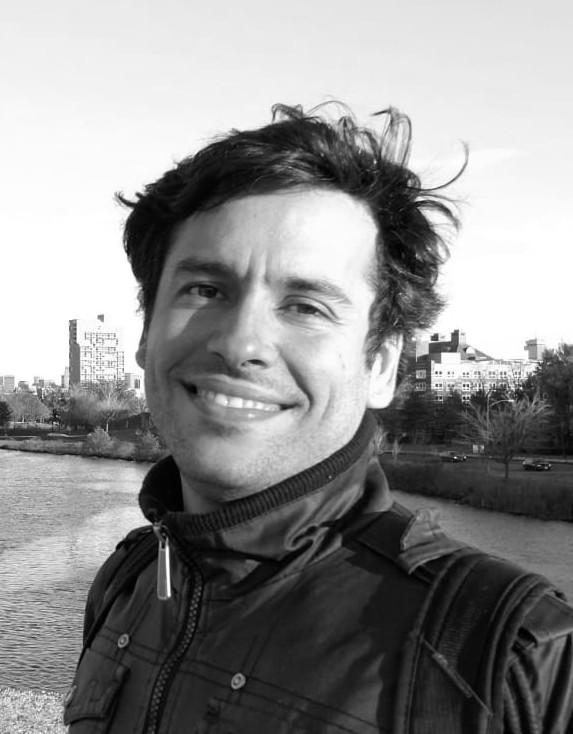}}] {Tales Imbiriba} (Member, IEEE)  is an Assistant Professor in the Computer Science Department at the University of Massachusetts Boston (UMB), Boston MA. Prior to joining UMB, he served as a Research Professor at the ECE dept., and Senior Research Scientist at the Institute for Experiential AI, both at Northeastern University (NU), Boston, MA, USA. He
received his Doctorate degree from the Department of Electrical Engineering (DEE) of the Federal University of Santa Catarina (UFSC), Florian\'opolis, Brazil, in 2016. He served as a Postdoctoral Researcher at the DEE--UFSC (2017--2019) and at the ECE dept. of the NU (2019--2021). 
His research focuses on Bayesian inference, online learning, and physics-guided machine learning, with applications in human-centered technologies, remote sensing of the environment, and enhanced security technologies.
\end{IEEEbiography}

\begin{IEEEbiography}[{\includegraphics[width=1in,height=1.25in,clip,keepaspectratio]{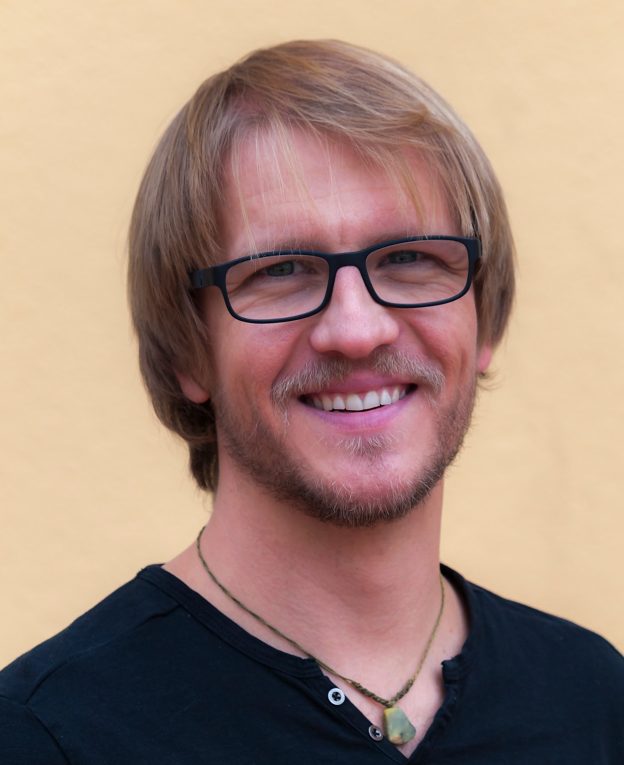}}]{Pau Closas}(Senior Member, IEEE),
is an Associate Professor in Electrical and Computer Engineering at Northeastern University, Boston MA.
He received the MS and PhD in Electrical Engineering from UPC in 2003 and 2009, respectively. He also holds a MS in Advanced Maths and Mathematical Engineering from UPC since 2014. He is the recipient of the EURASIP Best PhD Thesis Award 2014, the $9^{th}$ Duran Farell Award for Technology Research, the $2016$ ION's Early Achievements Award, $2019$ NSF CAREER Award, and the IEEE AESS Harry Rowe Mimno Award in $2022$. His primary areas of interest include statistical signal processing, stochastic filtering, robust filtering, and machine learning, with applications to positioning and localization systems. He is EiC for the IEEE AESS Magazine and volunteered in multiple editorial roles (e.g. NAVIGATION, Proc. IEEE, IEEE Trans. Veh. Tech., and IEEE Sig. Process. Mag.), and has been actively involved in organizing committees of a number of conference such as EUSIPCO (2011, 2019, 2021, 2022), IEEE SSP'16, IEEE/ION PLANS (2020, 2023), or IEEE ICASSP'20.
\end{IEEEbiography}



\newpage
\appendix




\subsection{Algorithms}\label{ap:algorithms}
In this appendix, we bring the pseudo-code for the conceptual BCFL algorithm~\ref{alg:BCFL}, which retains all possible associations at each iteration, and the approximations with pseudo-codes given in Algorithm~\ref{alg:BCFL-G} for BCFL-G and Algorithm~\ref{alg:BCFL-approx} for both BCFL-C and BCFL-MH. 

Note that in the conceptual algorithm~\ref{alg:BCFL}, there is no need for identifying specific subsets of associations, instead, all associations are retained. This approach allows for the simultaneous update of association weights and the local posterior. Conversely, the approximated algorithm requires a pre-selection of associations to be preserved, necessitating a decision prior to local model training. Consequently, the process involves two communication rounds: the first to upload association weights and confirm the decision, followed by a second where the decision is downloaded locally to guide training, thereby reducing superfluous computational expenses.

For algorithm BCFL-G~\ref{alg:BCFL-G}, $M_{\text{max}}=1$, the generated hypothesis will be singular, significantly simplifying computations. Moreover, efficiency gains can be achieved by localizing decision-making, which eliminates the need for one round of communication since the optimal association is selected through a greedy approach known locally, removing the necessity to transmit data back to the server. This process aligns with the IFCA method introduced by \cite{ghosh2020efficient}, as detailed in Algorithm \ref{alg:BCFL-G}. 

Regarding the other two approximations BCFL-C and BCFL-MH, see Algorithm \ref{alg:BCFL-approx}, the complexity is mitigated by keeping only the most promising $M_{\text{max}}$ hypotheses under the assumption that, in each hypothesis, one client can only be associated to one cluster. If all updated hypotheses are merged into a single one for the next iteration, we call it BCFL-C, whereas if we keep $M_{\text{max}}$ for the next iteration, it is referred to as BCFL-MH, as described in Algorithm \ref{alg:BCFL-approx}.

In the case of the BCFL-C, where $M_{\text{max}}$ is not equal to $1$, the server's role is to merge all $M_{\text{max}}$ hypotheses. Detailed information on how this merging is executed can be found in the Appendix~\ref{ap:Gaussian_fusion}.

As for the BCFL-MH algorithm, where $M_{\text{max}}$ are not equal to $1$, pruning is performed maintaining $M_{\text{max}}$ hypotheses with largest weights. 

\textbf{Communication and computation cost.}
Federated Learning incurs notable communication costs due to the regular transmission of model updates between numerous decentralized clients and a central server, with costs influenced by factors such as the number of clients, model size, update frequency, data distribution, channel quality, and so on. In our communication analysis within this study, we focus exclusively on quantifying the volume of parameters that must be transmitted during each communication round. As previously addressed, the association weights are transmitted initially, with the quantity of weights sent per round being $K C M_{\text{max}}$. In the case of BCFL-G, this transmission round is omitted, whereas for BCFL-C, it entails $K C$. Regarding the transmission of model parameters, the requisite size is $m K M_{\text{max}}$, where $m$ represents the size of an individual model. Notably, BCFL-MH incurs a significantly higher computation cost, amounting to $M_{\text{max}}$ times that of other methods.

Additionally, the BCFL-MH algorithm incurs a higher computational cost during local training compared to alternative methods. Consequently, its practicality in distributed systems that prioritize efficiency is limited. However, as demonstrated by the results, BCFL-MH is capable of outperforming its counterparts. Therefore, distributed systems where computational cost is not a primary concern could leverage BCFL-MH to enhance performance, which is a significant consideration. When efficiency is of greater importance, a greedy and consensus approach may be adopted to balance the trade-off between performance and computational expense.

In order to find the best $M_{\max}$ hypotheses, it is usually computationally intense if we do the global search, the complexity is usually known to be $\mathcal{O}(n^3)$ \cite{talbi2009metaheuristics}. However, in practice, finding the $M_{\max}$ can be solved in linear time in our case given that there is no constraint for a cluster to associate to multiple clients as the generic Hungarian algorithm imposes. Firstly, we identify the optimal association by finding the minimal value of each client, for which we can refer to the example in figure~\ref{fig:example of assoication}. This complexity is $\mathcal{O}(CK)$. Then, we employ heuristic methods to identify several sub-optimal paths (or associations) \cite{talbi2009metaheuristics}. Our approach modifies the optimal trajectory path by changing one index at a time, substituting it with another index that yields a minimal cost deviation from the optimal path. Repeating this process $M$ times results in a computational complexity of $\mathcal{O}(MCK)$, providing a feasible approach to approximating optimal solutions without the exhaustive search's computational burden. 




\begin{figure}[htbp]
\centering
\caption{Here we show a simple example of the assignment problem. The example of association with 2 clusters and 3 clients. The table is the loss of assignment from $C^j$ to $S^i$. The orange numbers are the optimal assignment loss. The optimal cost as the equation shows.}
\label{fig:example of assoication}

\begin{tabular}{c|cc}
Client $\backslash$ Model & $S^1$ & $S^2$ \\ \hline
$C^1$ & \textcolor{orange}{5} & 8 \\
$C^2$ & 8 & \textcolor{orange}{2} \\
$C^3$ & \textcolor{orange}{4} & 8 \\
\end{tabular}
\label{table:page99}

\vspace{5mm}  
\begin{align}\label{eq:example_assignment}
\text{Cost} &= \textrm{Tr}(A^\top L) = \textrm{Tr} \left( 
\begin{bmatrix}
1 & 0 \\
0 & 1 \\
1 & 0
\end{bmatrix}
\begin{bmatrix}
5 & 8 & 4 \\
8 & 2 & 8
\end{bmatrix}
\right) \nonumber \\
&= 5 + 3 + 4 = 11
\end{align}

\end{figure}

\subsection{$\mathsf{MERGE}(\cdot)$ operator: fusion of Gaussian densities}\label{ap:Gaussian_fusion}

As the $\mathsf{MERGE}$ operator used in BCFL-C algorithm we considered the arithmetic averaging (AA) aggregation approach discussed in \cite{li2019second}. Thus, given a Gaussian mixture $p(x)=\sum_{j=1}^M\phi_j\mathcal{N}(x; m_j, S_j)$, with weights $\phi_j$, means $m_j$ and covariances $S_j$, the application of the $\mathsf{MERGE}$ operator returns a single Gaussian $\mathcal{N}(x; \widetilde{m}, \widetilde{S})$ whose parameters are given as: 
\begin{align}\label{eq:JPDA_AA}
    \widetilde{m} &= \sum_{j=1}^M \phi_j m_j, \\
     \widetilde{S} &= \sum_{j=1}^M \phi_j (S_j +m_j m_j^\top - \widetilde{m}\widetilde{m}^\top).
\end{align}

\subsection{Additional experimental details}\label{app:fexps}

\subsubsection{Datasets}\label{app:datasets}
To construct our experimental scenarios we leverage four popular datasets from which we construct two feature- and two label-skewed experiments. 
The datasets are: 
\begin{enumerate}
    \item \textbf{Digits-Five} \citep{peng2019moment} consists of a collection of five popular digit datasets: MNIST (mt) ($55000$ samples), MNIST-M (mm) ($55000$ samples), Synthetic Digits (syn) ($25000$ samples), SVHN (sv)($73257$ samples), and USPS (up) ($7438$ samples). Each digit dataset includes a different style of 0-9 digit images. 
    \item \textbf{AmazonReview} \citep{blitzer2007biographies} AmazonReview is a dataset to tackle the task of identifying whether the sentiment of a product review is positive or negative. This dataset includes reviews from four different merchandise categories: Books (B) ($2834$ samples), DVDs (D) ($1199$ samples), Electronics (E) ($1883$ samples), and Kitchen and housewares (K) ($1755$ samples). 
    \item \textbf{Fashion-MNIST} \citep{xiao2017fashion} consists of $70000$ $28\times 28$ grayscale images in $10$ classes, with $60000$ training images and $10000$ test images.
    \item  \textbf{CIFAR-10} \citep{krizhevsky2009learning} provides $60000$ $32 \times 32$ color images in $10$ classes, with $6000$ images per class.
\end{enumerate}

\subsubsection{Details of experimental settings}\label{app:experiments}

\paragraph{Model related settings.} The models trained with the different FL methods in the experiments are neural networks. Particularly, we use small Convolutional Neural Networks (CNNs) with $3$ convolutional layers for the Digits-Five dataset; $3$ layers fully-connected layers for the AmazonReview dataset; and $2$ convolutional layers for both Fashion-MNIST and CIFAR-10 datasets. More details refer to Tables \ref{tb:digit5_str}--\ref{tb:cifar_str}. The optimization of the CNN was done using SGD with a learning rate $0.005$ and momentum $0.9$, with a batch size of $32$. For the training, we run $100$ global communication rounds, and the local steps in each communication are $10$. The warm-up steps is set to $2$.

\begin{table}[]
\centering
\caption{Detailed information of CNN for Digits-Five.}
\begin{tabular}{c|l}
\hline
\textbf{Layers}     & \multicolumn{1}{c}{\textbf{Details}}                                                             \\ \hline
Convolution & \begin{tabular}[c]{@{}l@{}}Conv2d(3, 64, kernel size = (5, 5), padding = 2)\\ BatchNorm2d(64)\\ ReLU()\\ MaxPool2d(3, 3)\end{tabular}  \\ \hline
Convolution & \begin{tabular}[c]{@{}l@{}}Conv2d(64, 64, kernel size = (5, 5), padding = 2)\\ BatchNorm2d(64)\\ ReLU()\\ MaxPool2d(3, 3)\end{tabular} \\ \hline
Convolution & \begin{tabular}[c]{@{}l@{}}Conv2d(64, 128, kernel size = (5, 5), padding = 2)\\ BatchNorm2d(128)\\ ReLU()\end{tabular}                 \\ \hline
Linear     & \begin{tabular}[c]{@{}l@{}}Linear(6272, 3072)\\ BatchNorm1d(3072)\\ ReLU()\end{tabular} \\ \hline
Linear     & \begin{tabular}[c]{@{}l@{}}Linear(3072, 2048)\\ BatchNorm1d(2048)\\ ReLU()\end{tabular} \\ \hline
Classifier & Linear(2048, 10)                                                                        \\ \hline
Loss       & CrossEntropy()                                                                          \\ \hline
\end{tabular}
\label{tb:digit5_str}
\end{table}

\begin{table}[]
\centering
\caption{Detailed information of CNN for AmazonReview.}
\begin{tabular}{c|l}
\hline
\textbf{Layers}                          & \multicolumn{1}{c}{\textbf{Details}}                                         \\ \hline
Linear                          & \begin{tabular}[c]{@{}l@{}}Linear(5000, 1000)\\ ReLU()\end{tabular} \\ \hline
Linear                          & \begin{tabular}[c]{@{}l@{}}Linear(1000, 500)\\ ReLU()\end{tabular}  \\ \hline
Linear                          & Linear(500, 100)                                                    \\ \hline
\multicolumn{1}{l|}{Classifier} & Linear(100, 2)                                                      \\ \hline
Loss                            & CrossEntropy()                                                      \\ \hline
\end{tabular}
\label{tb:amazon_str}
\end{table}

\begin{table}[]
\centering
\caption{Detailed information of CNN for Fashion-MNIST.}
\begin{tabular}{c|l}
\hline
\textbf{Layers}     & \multicolumn{1}{c}{\textbf{Details}} \\ \hline
Convolution & \begin{tabular}[c]{@{}l@{}}Conv2d(1, 16, kernel size = (5, 5), padding = 2)\\ BatchNorm2d(16)\\ ReLU()\\ MaxPool2d(2, 2)\end{tabular}    \\ \hline
Convolution & \begin{tabular}[c]{@{}l@{}}Conv2d(16, 32,  kernel size = (5, 5),  padding = 2)\\ BatchNorm2d(16)\\ ReLU()\\ MaxPool2d(2, 2)\end{tabular} \\ \hline
Classifier & Linear(7 $\cdot$ 7 $\cdot$ 32, 10)          \\ \hline
Loss       & CrossEntropy()              \\ \hline
\end{tabular}
\label{tb:fash_str}
\end{table}

\begin{table}[]
\centering
\caption{Detailed information of CNN for CIFAR-10.}
\begin{tabular}{c|l}
\hline
\textbf{Layers}      & \multicolumn{1}{c}{\textbf{Details}}                                                                           \\ \hline
Convolution & \begin{tabular}[c]{@{}l@{}}Conv2d(3, 6, kernel size = (5, 5))\\ ReLU()\\ MaxPool2d(2, 2)\end{tabular} \\ \hline
Convolution & \begin{tabular}[c]{@{}l@{}}Conv2d(6, 6, kernel size = (5, 5))\\ ReLU()\\ MaxPool2d(2, 2)\end{tabular} \\ \hline
Linear      & \begin{tabular}[c]{@{}l@{}}Linear(400,120)\\ ReLU()\end{tabular}                                      \\ \hline
Linear      & \begin{tabular}[c]{@{}l@{}}Linear(120,84)\\ ReLU()\end{tabular}                                       \\ \hline
Classifier  & Linear(84, 10)                                                                                        \\ \hline
Loss        & CrossEntropy()                                                                                        \\ \hline
\end{tabular}
\label{tb:cifar_str}
\end{table}

\paragraph{FL settings} 
%
To simulate label distribution skewness across clients, we use a method based on the use of a Dirichlet distribution for sampling labels, method that has been used in many recent FL studies \citep{li2021federated}.
Specifically, for a given client $j$, we define the probability of sampling data from the $k \in \{1,\dots,l\}$ label as the vector $(p_{j,1},\dots,p_{j,l}) \sim \operatorname{Dir}(\boldsymbol{\alpha})$, where $\operatorname{Dir}(\cdot)$ denotes the Dirichlet distribution and $\boldsymbol{\alpha} = (\alpha_{j,1}, \dots, \alpha_{j,l})$ is the concentration vector parameter ($\alpha_{j,k}>0$, $\forall k$). 
The advantage of this approach is that the imbalance level can be flexibly changed by adjusting the concentration parameter $\alpha_{j,k}$. If it is set to a smaller value, the partition is more unbalanced.

In our work, to generate our label-skewed dataset, we follow a two-step process as in \cite{ma2022convergence}. First, we divide the dataset into four groups, with a common concentration parameter of $ \alpha=0.1 $ for all the labels. This ensures that each group's distribution is different from the others. Then, we further divide the data into 10 clients per group, with a concentration parameter of $ \alpha =10 $, to control that the clients from the same group have a similar distribution and thus could be clustered together.

\paragraph{Hardware settings} 
For our experiments, we primarily used the V100 GPU with 32GB of memory to obtain the main results. Initial experiments were conducted on a Mac M1 Pro with 16GB of RAM, which is also sufficient for some tasks. Most algorithms complete execution in under an hour. However, for BCFL-MH, the computational cost is significantly higher, often requiring several hours, especially when $M=6$.

\subsubsection{Additional Experimental results.}\label{ap:moreresults}

\begin{table*}[htb]
\caption{Performance comparison on cluster-wise non-IID. For Digits-Five and AmazonReview feature-skewed data scenarios, Feature $(K, C/K)$ indicates $K$ data groups with $C/K$ clients per group. For Fashion-MNIST and CIFAR-10 label-skewed data scenarios, Label $(K, C/K, \alpha)$ indicates 
also the value of the Dirichlet concentration parameter $\alpha$.}
\centering
\begin{tabular}{l|cc|cc|cc|cc}
\toprule
\multicolumn{1}{c|}{Datasets} &
  \multicolumn{2}{c}{Digits-Five} &
  \multicolumn{2}{c}{AmazonReview} &
  \multicolumn{2}{c}{Fashion-MNIST} &
  \multicolumn{2}{c}{CIFAR-10} \\ \midrule
\multicolumn{1}{l|}{non-IID setting} &
  \multicolumn{2}{c}{Feature $(5, 2)$} &
  \multicolumn{2}{c}{Feature $(4, 2)$} &
  \multicolumn{2}{c}{Label $(4,10, 0.1)$} &
  \multicolumn{2}{c}{Label $(4,10, 0.1)$} \\ \midrule
Methods &
  \multicolumn{1}{c}{Acc} &
  \multicolumn{1}{c}{F1} &
  \multicolumn{1}{c}{Acc} &
  \multicolumn{1}{c}{F1} &
  \multicolumn{1}{c}{Acc} &
  \multicolumn{1}{c}{F1} &
  \multicolumn{1}{c}{Acc} &
  \multicolumn{1}{c}{F1} \\ \bottomrule\toprule
BCFL-G &
  $94.35$ &
  $94.16$ &
  $87.53$ &
  $87.5$ &
  $96.81$ &
  $93.51$ &
  $64.73$ &
  $44.22$ \\ 
BCFL-G-W &
  $94.48$ &
  $94.29$ &
  $87.58$ &
  $87.57$ &
  $96.69$ &
  $92.21$ &
  $72.95$ &
  $52.01$ \\ \midrule
BCFL-C-3 &
  $94.04$ &
  $93.84$ &
  $87.95$ &
  $87.93$ &
  $96.84$ &
  $90.16$ &
  $72.12$ &
  $50.97$ \\ 
BCFL-C-3-W &
  $94.35$ &
  $94.17$ &
  $88.22$ &
  $88.20$ &
  $96.88$ &
  $90.06$ &
  $73.18$ &
  $52.61$ \\ 
BCFL-C-6 &
  $94.14$ &
  $93.93$ &
  $88.11$ &
  $88.09$ &
  $96.58$ &
  $86.55$ &
  $68.73$ &
  $47.40$ \\ 
BCFL-C-6-W &
  $94.42$ &
  $94.24$ &
  $87.96$ &
  $87.95$ &
  $96.71$ &
  $87.31$ &
  $72.22$ &
  $50.02$ \\ \midrule
BCFL-MH-3 &
  $95.39$ &
  $95.22$ &
  $\textbf{88.74}$ &
  $\textbf{88.74}$ &
  $97.13$ &
  $\textbf{93.70}$ &
  $74.35$ &
  $\textbf{56.24}$ \\ 
BCFL-MH-3-W &
  $95.83$ &
  $95.70$ &
  $88.38$ &
  $88.38$ &
  $\textbf{97.40}$ &
  $93.42$ &
  $76.19$ &
  $58.42$ \\ 
BCFL-MH-6 &
  $\textbf{96.02}$ &
  $\textbf{95.88}$ &
  $88.16$ &
  $88.15$ &
  $\textbf{97.27}$ &
  $92.56$ &
  $\textbf{75.26}$ &
   $53.45$ \\ 
BCFL-MH-6-W &
  $\textbf{96.22}$ &
  $\textbf{96.08}$ &
  $88.43$ &
  $88.43$ &
  $97.33$ &
  $90.62$ &
  $\textbf{77.56}$ &
  $\textbf{59.18}$ \\ 
  \bottomrule
\end{tabular}
\label{tb:table1_warmup}
\end{table*}

\paragraph{Warm-up comparison}
We conducted additional experiments beyond those reported in the main body of the paper. The results, shown in Table \ref{tb:table1_warmup}, compare algorithms with a warm-up setting designed to improve the initialization of the local models, potentially enhancing the overall FL solution. In this work, warm-up is implemented in two steps: first, using a Euclidean-distance metric~\cite{ma2022convergence} to cluster the parameters of local models, and then aggregating them into a merged model, which is shared among those in the same cluster. Experiments (denoted with a '-W' suffix to indicate the use of warm-up) reveal that while warm-up can be beneficial for convergence time and accuracy, the BCFL schemes demonstrate competitive performances without it, making warm-up not strictly necessary.

\begin{figure}
\centering
\includegraphics[width=0.24\textwidth]{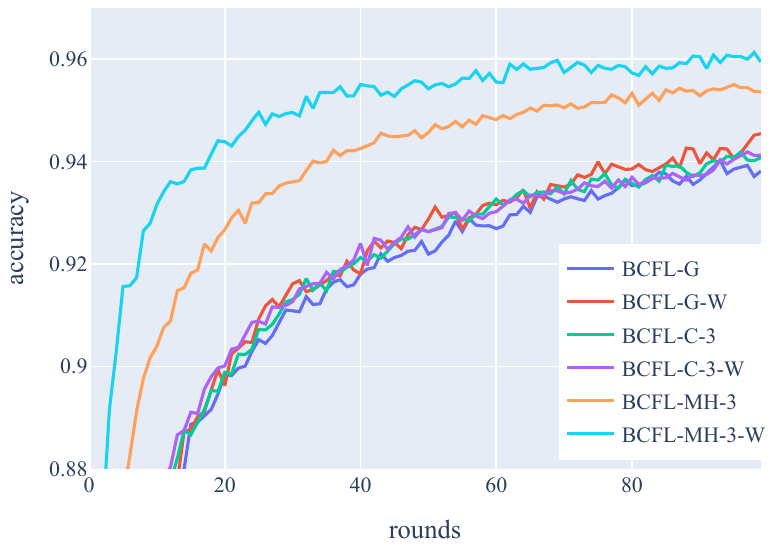}\hfill
\includegraphics[width=0.24\textwidth]{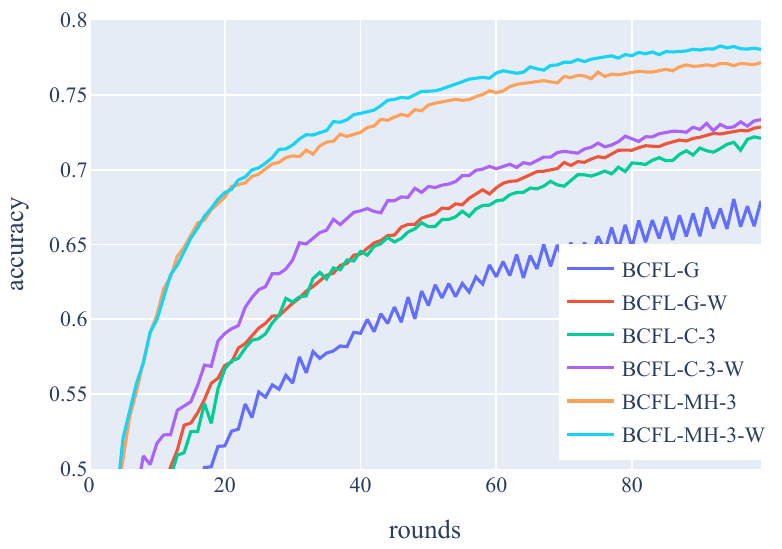}
\caption{Warm-up comparison of accuracy for Digits-Five (left panel) and CIFAR-10 (right panel)}
\label{fig:warm_up_accuracy_rounds}
\end{figure}

We evaluate the effect of using a warm-up stage for the BCFL methods. The convergence curves depicted in Figure~\ref{fig:warm_up_accuracy_rounds} indicate that warm-up can slightly accelerate convergence. Similar figures for the other datasets are provided in Appendix \ref{ap:moreresults}.

\begin{table}[]
\centering
\caption{Performance comparison.}
\begin{tabular}{|l|cccc|}
\hline
\multicolumn{1}{|c}{Datasets} & \multicolumn{2}{|c|}{Fashion-MNIST}                           & \multicolumn{2}{|c|}{AmazonReview}                            \\ \hline
Non-IID setting              & \multicolumn{2}{|c|}{Label $(0, 40, 0.1)$}                      & \multicolumn{2}{|c|}{Feature $(4, 10)$}                         \\ \hline
Methods                      & \multicolumn{1}{|l}{Accuracy} & \multicolumn{1}{l|}{F1} & \multicolumn{1}{l}{Accuracy} & \multicolumn{1}{l|}{F1} \\ \hline
FedAvg   & $88.64$          & $45.9$           & $87.86$          & 87.77         \\ \hline
WeCFL    & $90.67$         & $55.39$          & $85.63$          & $85.56$         \\ \hline
BCFL-G      & $92.79$          & $\textbf{69.13}$ & $88.01$          & 87.95         \\ \hline
BCFL-G-W    & $92.46$         & $62.75$          & $87.24$          & $87.17$         \\ \hline
BCFL-C-3   & $91.24$          & $61.27$         & $\textbf{88.36}$ & $\textbf{88.3}$ \\ \hline
BCFL-C-3-W & $92.39$          & $60.87$          & $87.38$         & $87.34$         \\ \hline
BCFL-C-6   & $92.17 $         & $60.76$          & $88.24$          & $88.18$         \\ \hline
BCFL-C-6-W & $90.92$          & $62.62$          & $86.86 $         & $86.79$         \\ \hline
BCFL-MH-3    & $93.41$          & $67.48$          & $87.21$          & 87.14         \\ \hline
BCFL-MH-3-W  & $93.49$          & $63.42$          & $87.12$          & $87.06$         \\ \hline
BCFL-MH-6    & $\textbf{94.29}$ & $68.69$          & $87.87$          & $87.81$         \\ \hline
BCFL-MH-6-W                      & $94.22$                        & $67.6$                         & $87.58$        & $87.48$         \\ \hline
\end{tabular}
\label{tb:table2}
\end{table}

\paragraph{Performance comparison with different client group setting}
We conducted additional experiments to those reported in the main body of the paper, the results of which are shown in Table \ref{tb:table2} for Fashion-MNIST and AmazonReview. These experiments consider different label- and feature-skewed configurations to those in the experiments of the main body of the paper. 
For Fashion-MNIST we considered the Label$(0,40,0.1)$ configuration, indicating no groups, $C=40$ clients, and $\alpha=0.1$.
%
For AmazonReview, we used a large number of clients ($40$) to showcase the results in larger datasets. In both cases, BCFL variants outperformed WeCFL and FedAvg. 


\begin{table*}[]
\centering
\caption{Fashion-MNIST comparison of different $K$ with non-IID data setting Label$(4,10,0.1)$}
\begin{tabular}{c|cc|cc|cc|cc|cc}
\hline
$K$ &
  \multicolumn{2}{c|}{2} &
  \multicolumn{2}{c|}{3} &
  \multicolumn{2}{c|}{4} &
  \multicolumn{2}{c|}{5} &
  \multicolumn{2}{c}{6} \\ \hline
Methods &
  Acc &
  F1 &
  Acc &
  F1 &
  Acc &
  F1 &
  Acc &
  F1 &
  Acc &
  F1 \\ \hline
FedAvg &
  {\color[HTML]{333333} 85.00} &
  {\color[HTML]{333333} 53.14} &
  {\color[HTML]{333333} 85.00} &
  {\color[HTML]{333333} 53.14} &
  {\color[HTML]{333333} 85.00} &
  {\color[HTML]{333333} 53.14} &
  {\color[HTML]{333333} 85.00} &
  {\color[HTML]{333333} 53.14} &
  {\color[HTML]{333333} 85.00} &
  53.14 \\ \hline
WeCFL &
  {\color[HTML]{333333} 91.79} &
  {\color[HTML]{333333} 75.99} &
  {\color[HTML]{333333} \textbf{96.74}} &
  {\color[HTML]{333333} 88.92} &
  {\color[HTML]{333333} \textbf{96.74}} &
  {\color[HTML]{333333} 92.00} &
  {\color[HTML]{333333} 96.65} &
  {\color[HTML]{333333} \textbf{92.21}} &
  {\color[HTML]{333333} 96.73} &
  {\color[HTML]{333333} 91.98} \\ \hline
BCFL-G &
  {\color[HTML]{333333} 88.98} &
  {\color[HTML]{333333} 71.78} &
  {\color[HTML]{333333} 89.40} &
  {\color[HTML]{333333} 72.10} &
  96.81 &
  \textbf{93.51} &
  \textbf{96.91} &
  93.19 &
  96.83 &
  92.98 \\ \hline
BCFL-C-3 &
  {\color[HTML]{333333} 91.84} &
  {\color[HTML]{333333} 72.05} &
  {\color[HTML]{333333} 95.82} &
  {\color[HTML]{333333} 83.15} &
  {\color[HTML]{333333} \textbf{96.84}} &
  {\color[HTML]{333333} 90.16} &
  {\color[HTML]{333333} 96.83} &
  {\color[HTML]{333333} \textbf{93.5}} &
  {\color[HTML]{333333} 96.76} &
  {\color[HTML]{333333} 92.44} \\
BCFL-C-6 &
  {\color[HTML]{333333} 92.87} &
  {\color[HTML]{333333} 71.25} &
  {\color[HTML]{333333} 96.89} &
  {\color[HTML]{333333} 90.51} &
  96.58 &
  {\color[HTML]{333333} 86.55} &
  {\color[HTML]{333333} \textbf{97.02}} &
  {\color[HTML]{333333} \textbf{93.22}} &
  {\color[HTML]{333333} 96.84} &
  92.87 \\ \hline
BCFL-MH-3 &
  {\color[HTML]{333333} 94.39} &
  {\color[HTML]{333333} 74.60} &
  {\color[HTML]{333333} 97.23} &
  {\color[HTML]{333333} 92.53} &
  {\color[HTML]{333333} 97.13} &
  {\color[HTML]{333333} 93.70} &
  {\color[HTML]{333333} \textbf{97.32}} &
  {\color[HTML]{333333} 93.18} &
  {\color[HTML]{333333} 97.35} &
  \textbf{94.02} \\
BCFL-MH-6 &
  {\color[HTML]{333333} 87.21} &
  {\color[HTML]{333333} 63.59} &
  93.42 &
  82.49 &
  {\color[HTML]{333333} \textbf{97.27}} &
  {\color[HTML]{333333} \textbf{94.56}} &
  {\color[HTML]{333333} 97.23} &
  {\color[HTML]{333333} 92.46} &
  {\color[HTML]{333333} \textbf{97.27}} &
  {\color[HTML]{333333} 94.16} \\ \hline
\end{tabular}
\label{tb:table_sens1}
\end{table*}

\begin{table*}[]
\centering
\caption{AmazonReview comparison of different $K$ with non-IID data setting Feature$(4,10)$}
\begin{tabular}{c|cc|cc|cc|cc|cc}
\hline
$K$ & \multicolumn{2}{c|}{2} & \multicolumn{2}{c|}{3} & \multicolumn{2}{c|}{4} & \multicolumn{2}{c|}{5} & \multicolumn{2}{c}{6} \\ \hline
Methods & Acc & F1 & Acc & F1 & Acc & F1 & Acc & F1 & Acc & F1 \\ \hline
FedAvg & 87.71 & 87.70 & 87.71 & 87.70 & 87.71 & 87.70 & 87.71 & 87.70 & 87.71 & 87.70 \\ \hline
WeCFL & 88.35 & \textbf{79.86} & \textbf{88.53} & 79.42 & 88.31 & 78.75 & 88.02 & 78.43 & 88.34 & 78.43 \\ \hline
BCFL-G & 87.68 & 80.03 & 89.00 & 80.25 & 87.53 & \textbf{87.50} & \textbf{89.30} & 83.4 & 89.27 & 81.92 \\ \hline
BCFL-C-3 & 88.55 & 80.19 & \textbf{89.31} & 80.98 & 89.19 & 82.11 & 89.07 & \textbf{82.44} & 89.26 & 80.61 \\
BCFL-C-6 & 88.50 & 80.86 & 88.64 & 81.48 & 88.66 & 81.63 & 88.76 & \textbf{82.60} & \textbf{88.90} & 81.25 \\ \hline
BCFL-MH-3 & 88.66 & 81.47 & \textbf{89.56} & 81.38 & 89.04 & 82.68 & 89.08 & 82.96 & 89.24 & \textbf{83.25} \\
BCFL-MH-6 & 88.08 & 80.80 & \textbf{88.90} & 80.95 & 88.65 & \textbf{82.78} & 88.84 & 82.64 & 88.77 & 82.47 \\ \hline
\end{tabular}
\label{tb:table_sens2}
\end{table*}

\paragraph{Sensitivity with $K$}
We conducted additional experiments to study the sensitivity of the methods to misspecification of the number of cluster $K$. In practice, selecting the correct number of clusters might not be possible in certain applications, while in others we might have access to such information. The results of the sensitivity analysis to $K$ are shown in Tables \ref{tb:table_sens1} and \ref{tb:table_sens2}, where the correct number of clusters is $K=4$. It can be seen that when the assumed $K$ is smaller than $4$, the results degrade to some degree for all methods, while not dramatically. On the other hand, when the cluster number is bigger than $4$, the results are usually very similar to the results under the correct $K$ specification. 
A large number of assumed clusters is therefore desirable from a performance perspective, although it comes at a higher computational cost. 
A small number of clusters usually means less computation cost, but may reduce the personalized features of BCFL. 
In the open literature, related works refer to this challenge. Some suggest choosing the number of clusters by running a small experiment with a few rounds \citep{long2022multi}, while others present methods found in most clustering algorithms such as the information criterion approach \citep{goutte2001feature,sugar2003finding}, which are not seen in FL clustering. This may raise great interest in how to apply them, as well as how to learn $K$ also from the perspective of data association theory that drives our work. 

\begin{table*}[htb]
\caption{Table \ref{tb:table1} with standard deviation}
\tiny
\centering
\begin{tabular}{l|cc|cc|cc|cc}
\toprule
\multicolumn{1}{c|}{Datasets} &
  \multicolumn{2}{c}{Digits-Five} &
  \multicolumn{2}{c}{AmazonReview} &
  \multicolumn{2}{c}{Fashion-MNIST} &
  \multicolumn{2}{c}{CIFAR-10} \\ \midrule
\multicolumn{1}{l|}{non-IID setting} &
  \multicolumn{2}{c}{Feature $(5, 2)$} &
  \multicolumn{2}{c}{Feature $(4, 2)$} &
  \multicolumn{2}{c}{Label $(4,10, 0.1)$} &
  \multicolumn{2}{c}{Label $(4,10, 0.1)$} \\ \midrule
Methods &
  \multicolumn{1}{c}{Acc} &
  \multicolumn{1}{c}{F1} &
  \multicolumn{1}{c}{Acc} &
  \multicolumn{1}{c}{F1} &
  \multicolumn{1}{c}{Acc} &
  \multicolumn{1}{c}{F1} &
  \multicolumn{1}{c}{Acc} &
  \multicolumn{1}{c}{F1} \\ \bottomrule\toprule
FedAvg &
  $93.18 \pm 0.02$&
  $92.95 \pm 0.02$&
  $87.71 \pm 0.18$ &
  $87.70 \pm 0.18$ &
  $85.00 \pm 0.24$&
  $53.14 \pm 0.32$&
  $39.89 \pm 0.18$&
  $21.19 \pm 0.18$\\ \midrule
WeCFL &
  $93.81 \pm 0.20$ &
  $93.58 \pm 0.20$ &
  $85.57 \pm 0.33$ &
  $85.51 \pm 0.34$ &
  $96.74 \pm 0.10$&
  $92.0 \pm 0.50$&
  $72.91 \pm 0.25$ &
  $51.69 \pm 0.41$ \\ \midrule
BCFL-G &
  $94.35 \pm 0.00$ &
  $94.16 \pm 0.00$ &
  $87.53 \pm 0.00$ &
  $87.50 \pm 0.00$ &
  $96.81 \pm 0.06$&
  $93.51 \pm 0.92$&
  $64.73 \pm 7.74$ &
  $44.22 \pm 7.00$ \\ 
BCFL-G-W &
  $94.48 \pm 0.00$ &
  $94.29 \pm 0.00$ &
  $87.58 \pm 0.00$ &
  $87.57 \pm 0.00$ &
  $96.69 \pm 0.05$&
  $92.21 \pm 0.15$&
  $72.95 \pm 0.29$ &
  $52.01 \pm 0.64$ \\ \midrule
BCFL-C-3 &
  $94.04 \pm 0.11$ &
  $93.84 \pm 0.10$ &
  $87.95 \pm 0.18$ &
  $87.93 \pm 0.18$ &
  $96.84 \pm 0.02$&
  $90.16 \pm 1.30$&
  $72.12 \pm 0.21$ &
  $50.97 \pm 0.51$ \\ 
BCFL-C-3-W &
  $94.35 \pm 0.05$ &
  $94.17 \pm 0.04$ &
  $88.22 \pm 0.26$ &
  $88.20 \pm 0.26$ &
  $96.88 \pm 0.07$&
  $90.06 \pm 1.12$&
  $73.18 \pm 0.29$ &
  $52.61 \pm 0.41$ \\ 
BCFL-C-6 &
  $94.14 \pm 0.11$ &
  $93.93 \pm 0.12$ &
  $88.11 \pm 0.42$ &
  $88.09 \pm 0.43$ &
  $96.58 \pm 1.60$&
  $86.55 \pm 1.24$&
  $68.73 \pm 1.11$ &
  $47.40 \pm 1.29$ \\ 
BCFL-C-6-W &
  $94.42 \pm 0.02$ &
  $94.24 \pm 0.02$ &
  $87.96 \pm 0.42$ &
  $87.95 \pm 0.43$ &
  $96.71 \pm 0.04$&
  $87.31 \pm 2.45$&
  $72.22 \pm 0.14$ &
  $50.02 \pm 0.47$\\ \midrule
BCFL-MH-3 &
  $95.39 \pm 0.32$ &
  $95.22 \pm 0.35$ &
  $\textbf{88.74} \pm 0.19$&
  $\textbf{88.74} \pm 0.19$&
  $97.13 \pm 0.33$&
  $\textbf{93.70} \pm 3.77$&
  $74.35 \pm 0.98$ &
  $56.24 \pm 2.69$ \\ 
BCFL-MH-3-W &
  $95.83 \pm 0.08$ &
  $95.70 \pm 0.09$ &
  $88.38 \pm 0.26$ &
  $88.38 \pm 0.26$ &
  $\textbf{97.40} \pm 0.14$&
  $93.42 \pm 0.83$&
  $76.19 \pm 0.33$ &
  $58.42 \pm 0.40$ \\ 
BCFL-MH-6 &
  $96.02 \pm 0.00$ &
  $95.88 \pm 0.00$ &
  $88.16 \pm 0.23$&
  $88.15 \pm 0.23$&
  $97.27 \pm 0.00$&
  $92.56 \pm 0.00$&
  $75.26 \pm 1.75$&
   $53.45 \pm 5.23$ \\ 
BCFL-MH-6-W &
  $\textbf{96.22} \pm 0.00$ &
  $\textbf{96.08} \pm 0.00$ &
  $88.43 \pm 0.18$&
  $88.43 \pm 0.18$&
  $97.33 \pm 0.07$&
  $90.62 \pm 1.01$&
  $\textbf{77.56} \pm 0.22 $ &
  $\textbf{59.18} \pm 0.49 $ \\ \bottomrule
\end{tabular}
\label{tb:table1_std}
\end{table*}

\begin{figure}
    \centering
    \begin{subfigure}[b]{0.24\textwidth}
        \includegraphics[width=\textwidth]{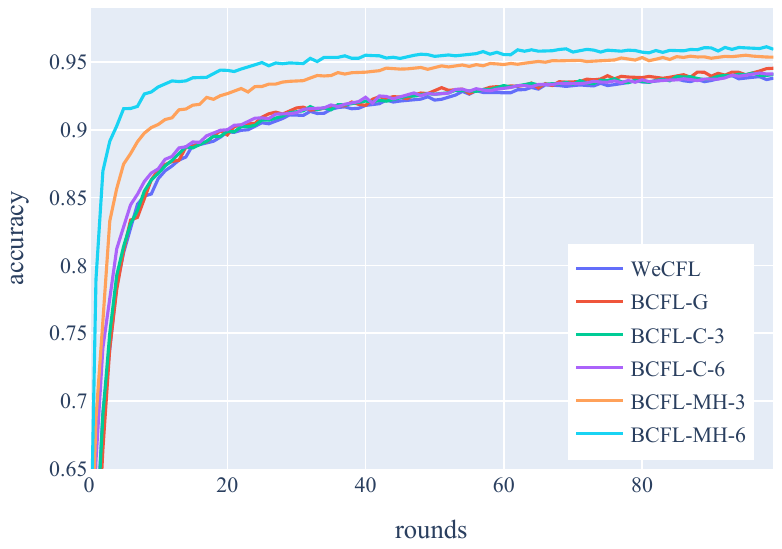}
        \caption{}
    \end{subfigure}
    \begin{subfigure}[b]{0.24\textwidth}
        \includegraphics[width=\textwidth]{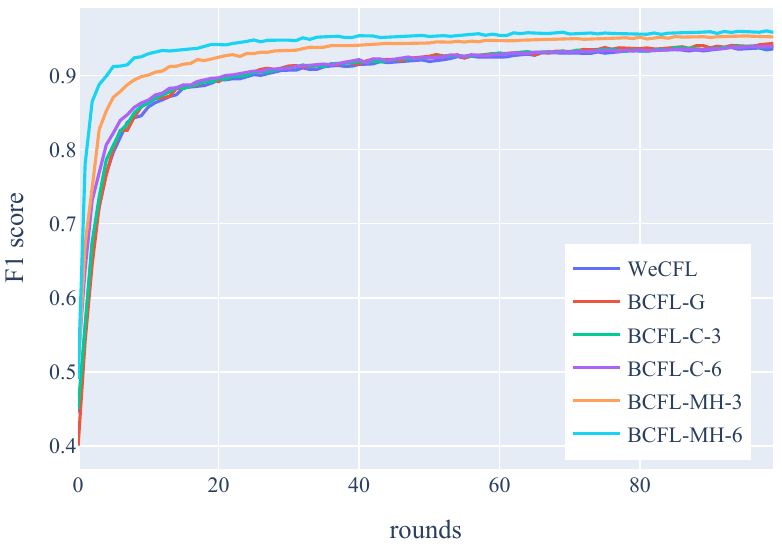}
        \caption{}
    \end{subfigure}
    \begin{subfigure}[b]{0.24\textwidth}
        \includegraphics[width=\textwidth]{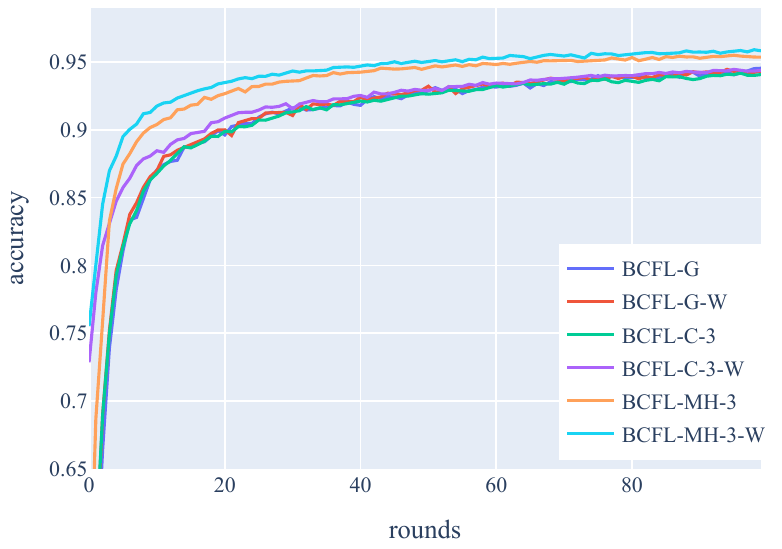}
        \caption{}
    \end{subfigure}
    \begin{subfigure}[b]{0.24\textwidth}
        \includegraphics[width=\textwidth]{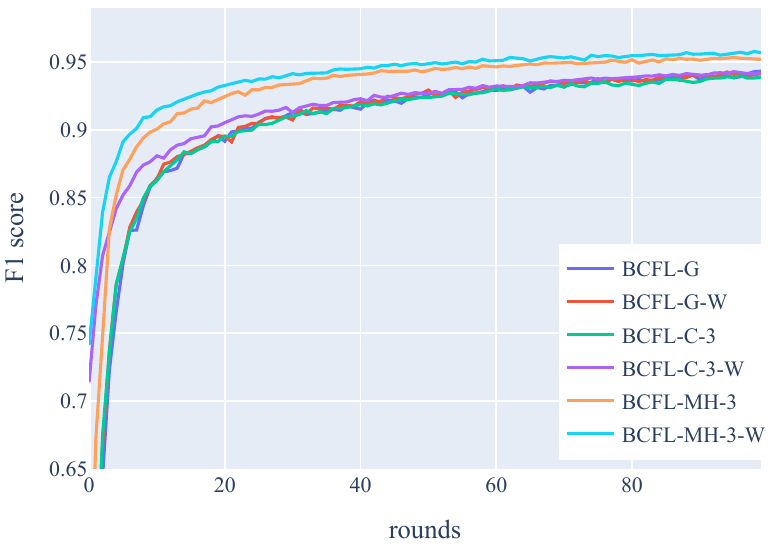}
        \caption{}
    \end{subfigure}
    \caption{Digits-Five: (a) Accuracy (b) F-1 score (c) Accuracy warm-up comparison (d) F-1 score warm-up comparison.}
    \label{fig:digit5_acc_f1}
\end{figure}

\begin{figure*}[htb]
    \centering
    \begin{subfigure}[b]{0.26\textwidth}
        \includegraphics[width=\textwidth]{figs/digit5_graph/node_wecflc.png}
        \caption{WeCFL}
    \end{subfigure}
    \hfill
    \begin{subfigure}[b]{0.26\textwidth}
        \includegraphics[width=\textwidth]{figs/digit5_graph/node_gnnc.png}
        \caption{BCFL-G}
    \end{subfigure}
    \hfill
    \begin{subfigure}[b]{0.26\textwidth}
        \includegraphics[width=\textwidth]{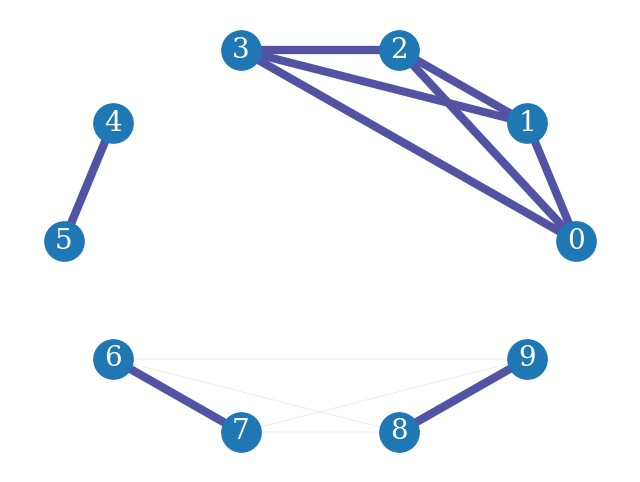}
        \caption{BCFL-G-W}
    \end{subfigure}
    \hfill
    \begin{subfigure}[b]{0.26\textwidth}
        \includegraphics[width=\textwidth]{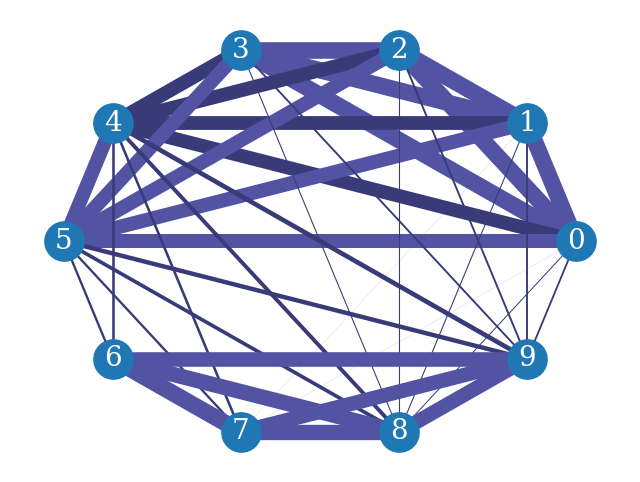}
        \caption{BCFL-C-3}
    \end{subfigure}
    \hfill
    \begin{subfigure}[b]{0.26\textwidth}
        \includegraphics[width=\textwidth]{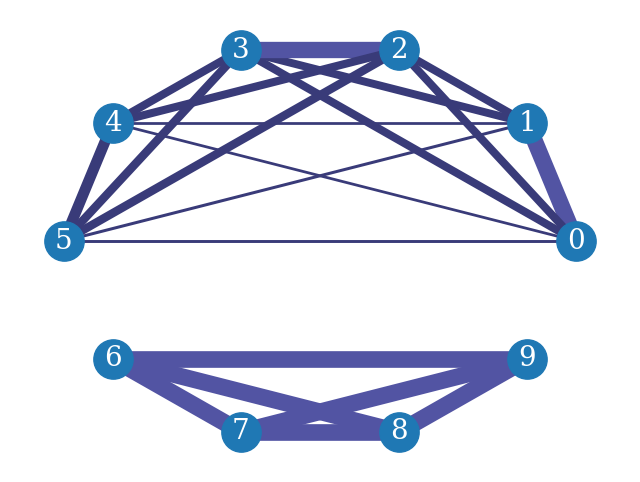}
        \caption{BCFL-C-3-W}
    \end{subfigure}
    \hfill
    \begin{subfigure}[b]{0.26\textwidth}
        \includegraphics[width=\textwidth]{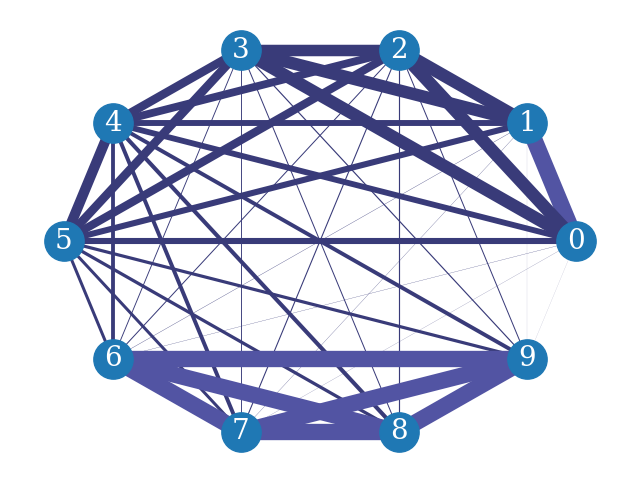}
        \caption{BCFL-C-6}
    \end{subfigure}
    \hfill
    \begin{subfigure}[b]{0.26\textwidth}
        \includegraphics[width=\textwidth]{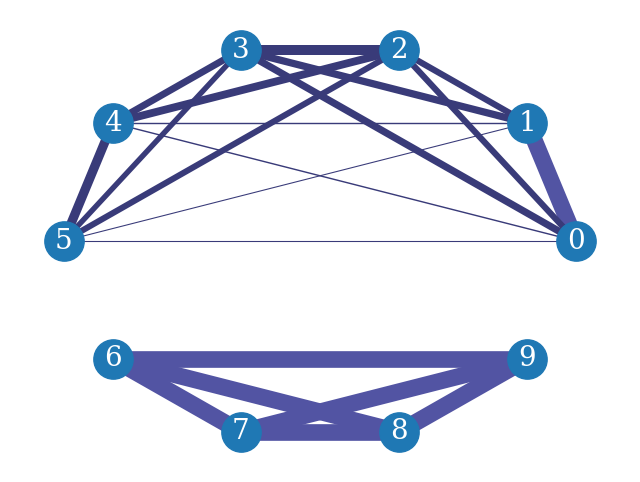}
        \caption{BCFL-C-6-W}
    \end{subfigure}
    \hfill
    \begin{subfigure}[b]{0.26\textwidth}
        \includegraphics[width=\textwidth]{figs/digit5_graph/node_mht3c.png}
        \caption{BCFL-MH-3}
    \end{subfigure}
    \hfill
    \begin{subfigure}[b]{0.26\textwidth}
        \includegraphics[width=\textwidth]{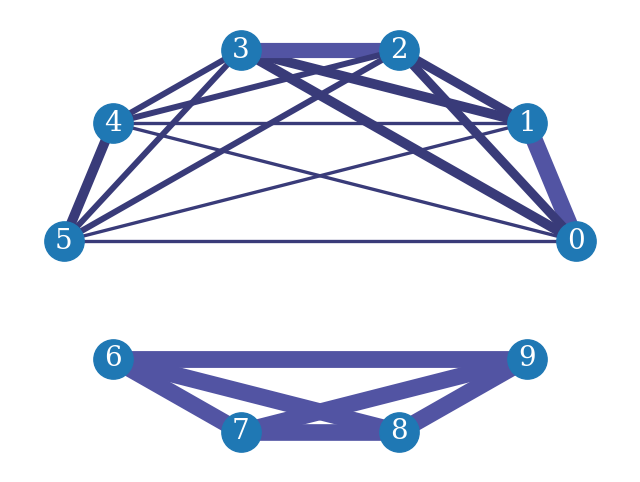}
        \caption{BCFL-MH-3-W}
    \end{subfigure}
    \hfill
    \begin{subfigure}[b]{0.26\textwidth}
        \includegraphics[width=\textwidth]{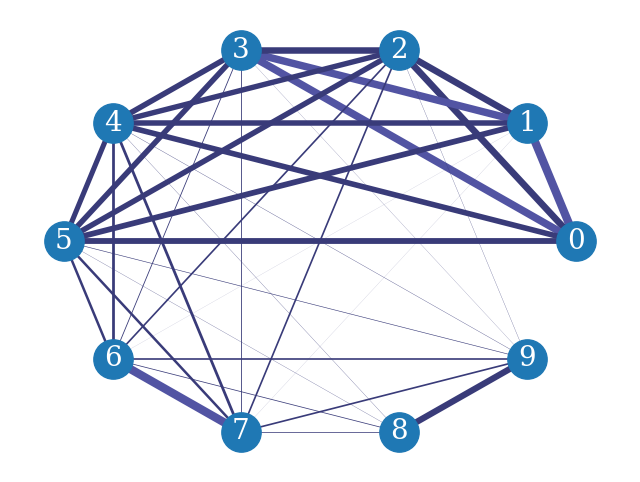}
        \caption{BCFL-MH-6}
    \end{subfigure}
    \hfill
    \begin{subfigure}[b]{0.26\textwidth}
        \includegraphics[width=\textwidth]{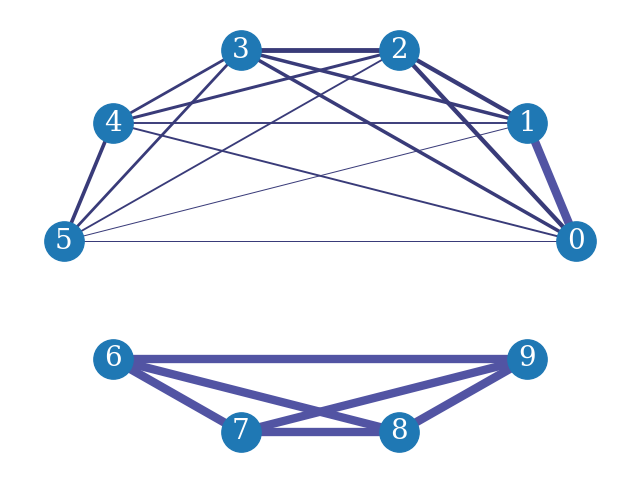}
        \caption{BCFL-MH-6-W}
    \end{subfigure}
    \caption{Client clustering during training for Digits-Five dataset. The experiment is such that of $K=5$ data groups with $C/K=2$ clients per group. It can be observed that the pairs of clients are grouped together, sometimes associated with other pairs as well.}
    \label{fig:cluster_digit5}
\end{figure*}

\begin{figure}
    \centering
    \begin{subfigure}[b]{0.24\textwidth}
        \includegraphics[width=\textwidth]{figs/amazon_acc/fig_acc_zon_829.pdf}
        \caption{}
    \end{subfigure}
    \begin{subfigure}[b]{0.24\textwidth}
        \includegraphics[width=\textwidth]{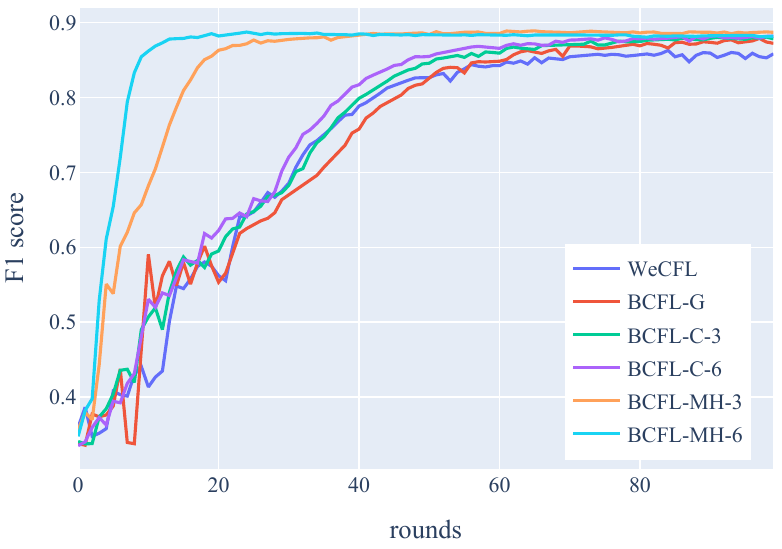}
        \caption{}
    \end{subfigure}
    \begin{subfigure}[b]{0.24\textwidth}
        \includegraphics[width=\textwidth]{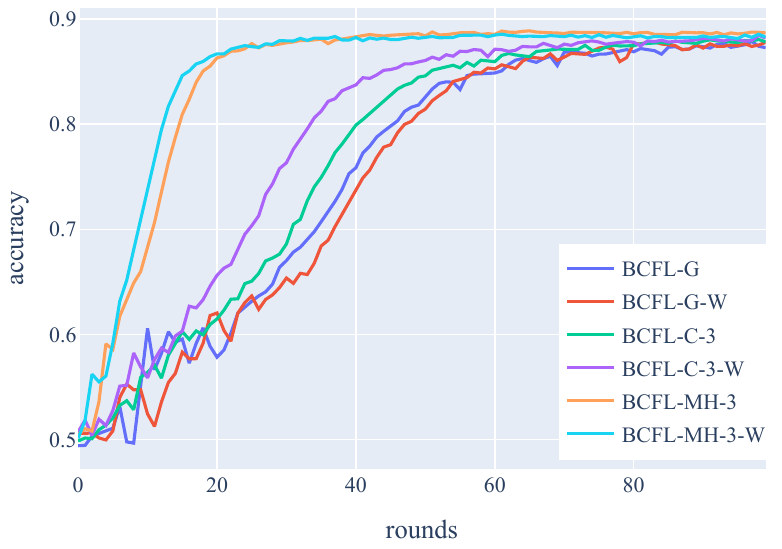}
        \caption{}
    \end{subfigure}
    \begin{subfigure}[b]{0.24\textwidth}
        \includegraphics[width=\textwidth]{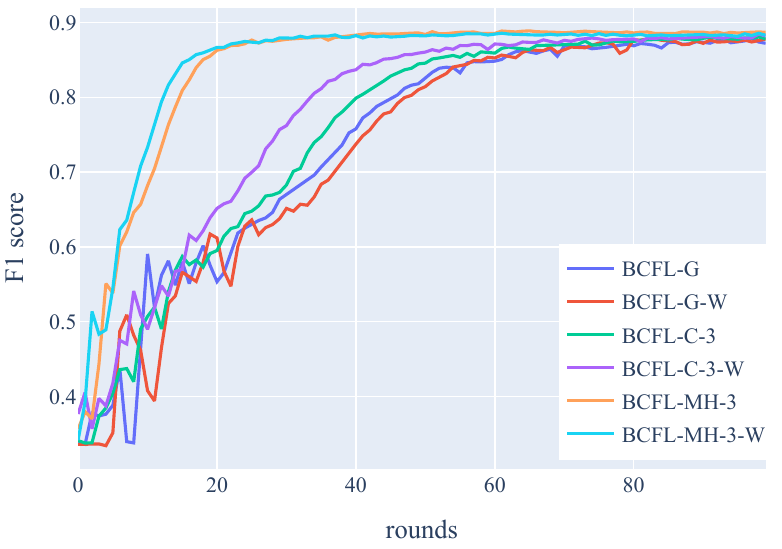}
        \caption{}
    \end{subfigure}
    \caption{AmazonReview: (a) Accuracy (b) F-1 score (c) Accuracy warm-up comparison (d) F-1 score warm-up comparison.}
    \label{fig:amazon_acc_f1}
\end{figure}

\begin{figure*}[htb]
    \centering
    \begin{subfigure}[b]{0.26\textwidth}
        \includegraphics[width=\textwidth]{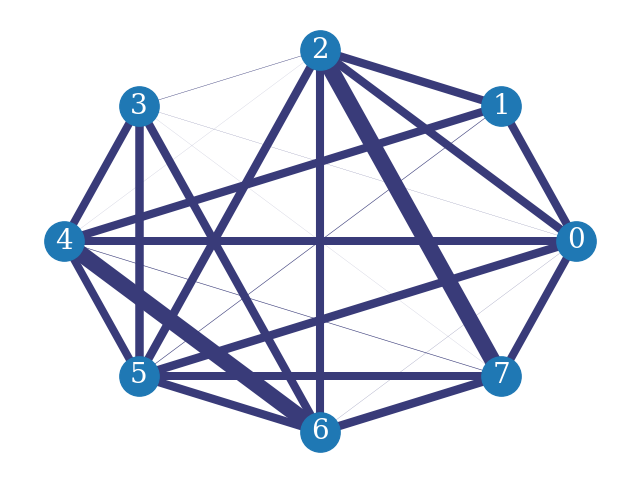}
        \caption{WeCFL}
    \end{subfigure}
    \hfill
    \begin{subfigure}[b]{0.26\textwidth}
        \includegraphics[width=\textwidth]{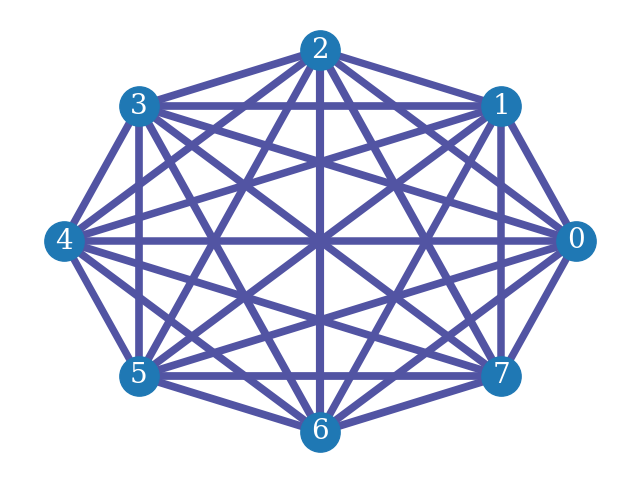}
        \caption{BCFL-G}
    \end{subfigure}
    \hfill
    \begin{subfigure}[b]{0.26\textwidth}
        \includegraphics[width=\textwidth]{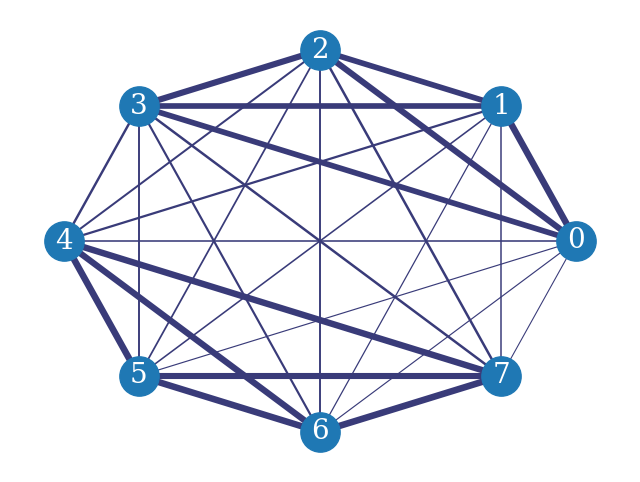}
        \caption{BCFL-G-W}
    \end{subfigure}
    \hfill
    \begin{subfigure}[b]{0.26\textwidth}
        \includegraphics[width=\textwidth]{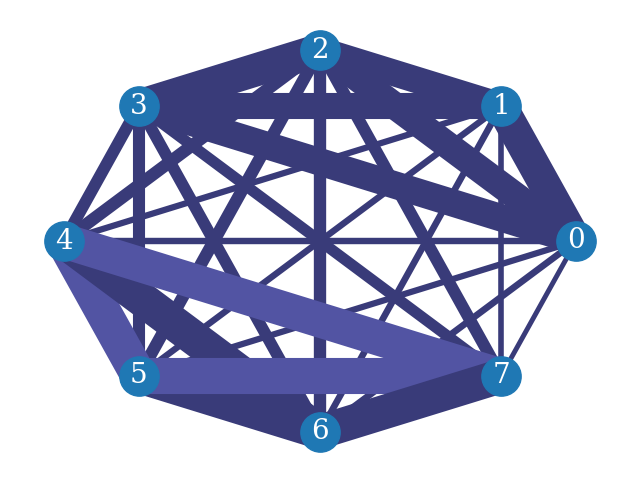}
        \caption{BCFL-C-3}
    \end{subfigure}
    \hfill
    \begin{subfigure}[b]{0.26\textwidth}
        \includegraphics[width=\textwidth]{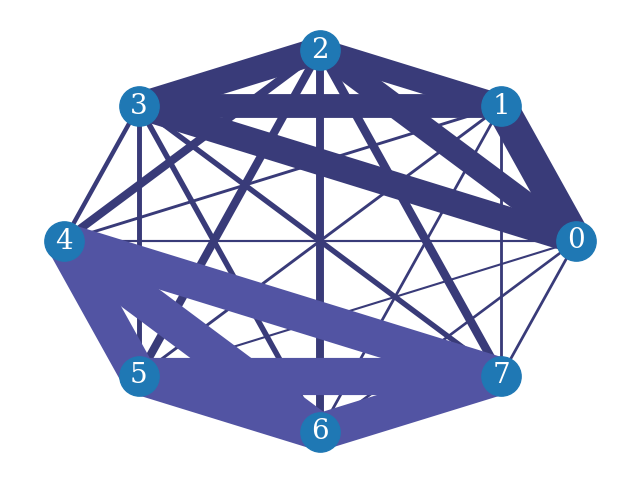}
        \caption{BCFL-C-3-W}
    \end{subfigure}
    \hfill
    \begin{subfigure}[b]{0.26\textwidth}
        \includegraphics[width=\textwidth]{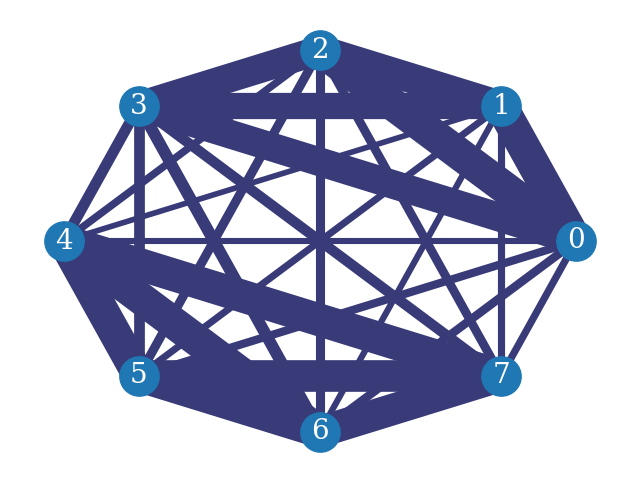}
        \caption{BCFL-C-6}
    \end{subfigure}
    \hfill
    \begin{subfigure}[b]{0.26\textwidth}
        \includegraphics[width=\textwidth]{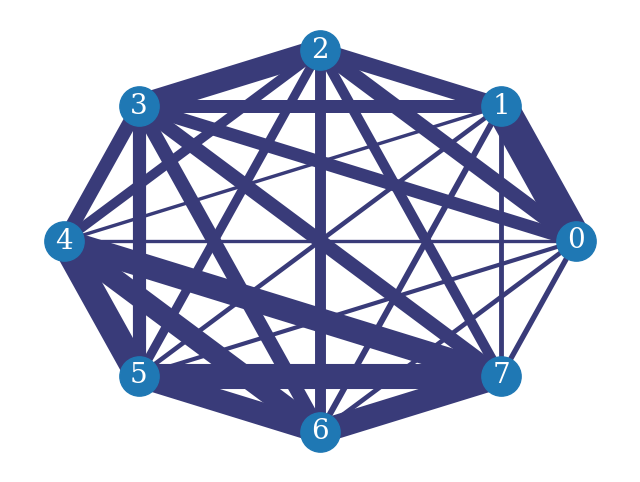}
        \caption{BCFL-C-6-W}
    \end{subfigure}
    \hfill
    \begin{subfigure}[b]{0.26\textwidth}
        \includegraphics[width=\textwidth]{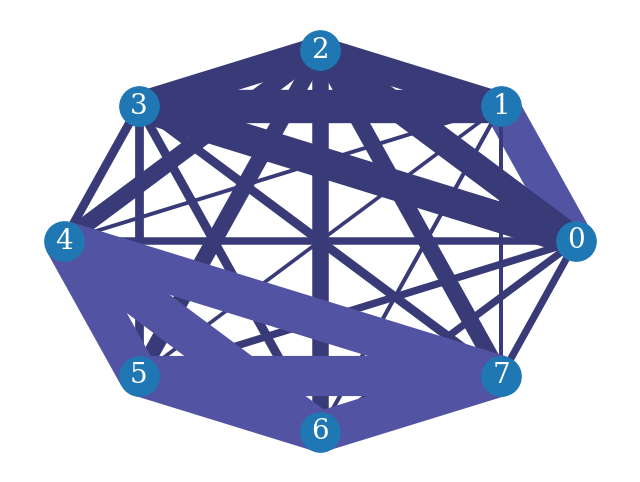}
        \caption{BCFL-MH-3}
    \end{subfigure}
    \hfill
    \begin{subfigure}[b]{0.26\textwidth}
        \includegraphics[width=\textwidth]{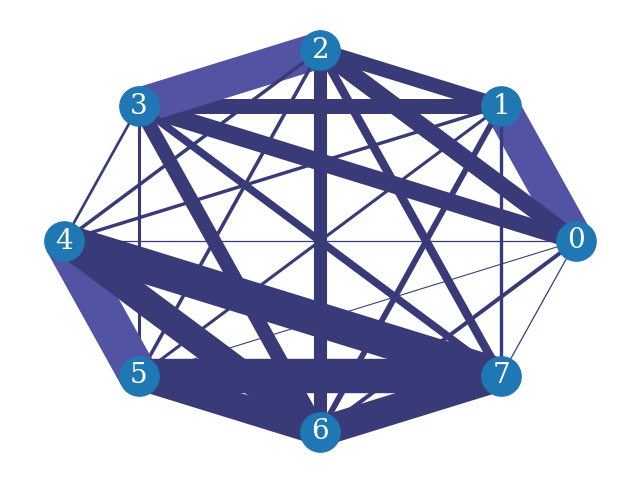}
        \caption{BCFL-MH-3-W}
    \end{subfigure}
    \hfill
    \begin{subfigure}[b]{0.26\textwidth}
        \includegraphics[width=\textwidth]{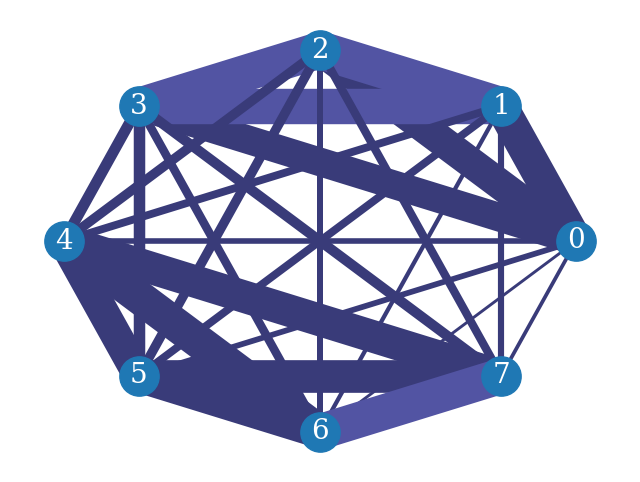}
        \caption{BCFL-MH-6}
    \end{subfigure}
    \hfill
    \begin{subfigure}[b]{0.26\textwidth}
        \includegraphics[width=\textwidth]{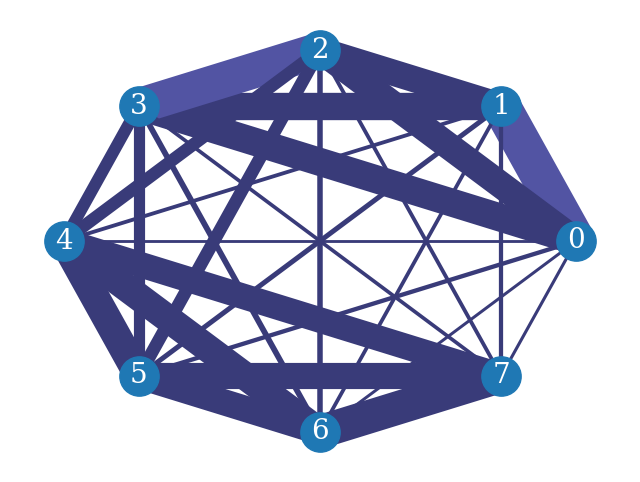}
        \caption{BCFL-MH-6-W}
    \end{subfigure}
    \caption{Client clustering during training for Amazon Review dataset. The experiment is such that of $K=4$ data groups with $C/K=2$ clients per group. It can be observed that the pairs of clients are grouped together, sometimes associated with other pairs as well.}
    \label{fig:cluster_amazon}
\end{figure*}

\begin{figure}
    \centering
    \begin{subfigure}[b]{0.24\textwidth}
        \includegraphics[width=\textwidth]{figs/fashion/fig_acc_og_fash.pdf}
        \caption{}
    \end{subfigure}
    \begin{subfigure}[b]{0.24\textwidth}
        \includegraphics[width=\textwidth]{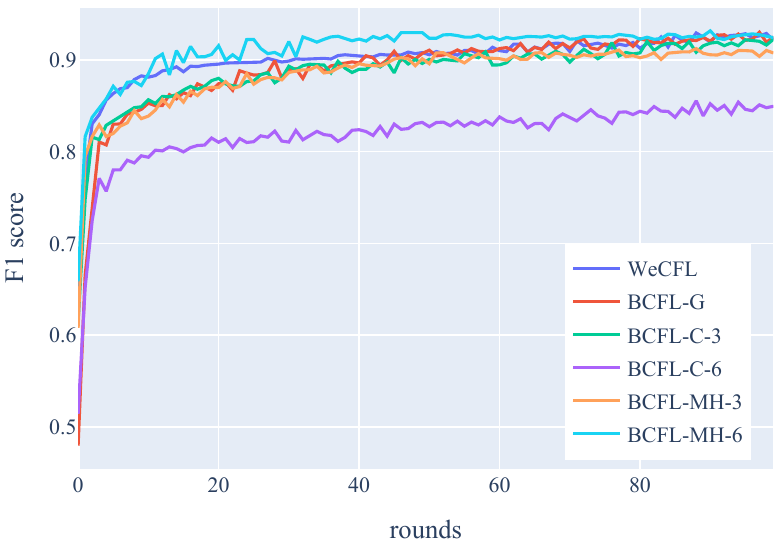}
        \caption{}
    \end{subfigure}
    \begin{subfigure}[b]{0.24\textwidth}
        \includegraphics[width=\textwidth]{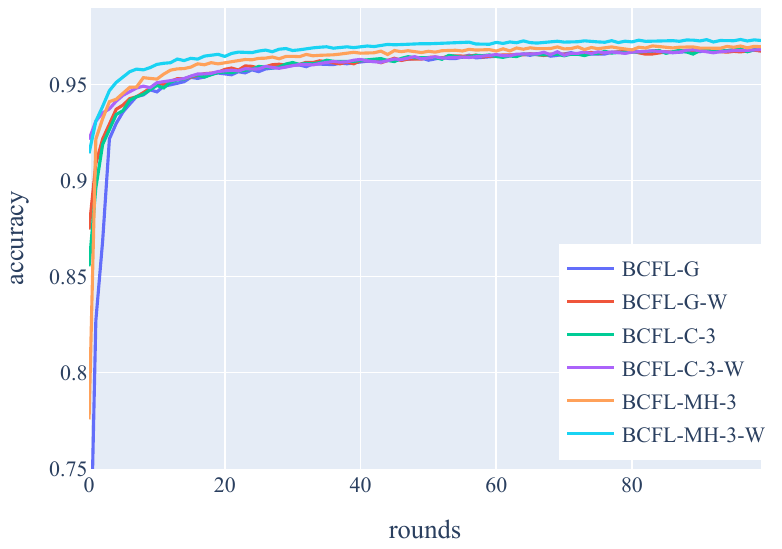}
        \caption{}
    \end{subfigure}
    \begin{subfigure}[b]{0.24\textwidth}
        \includegraphics[width=\textwidth]{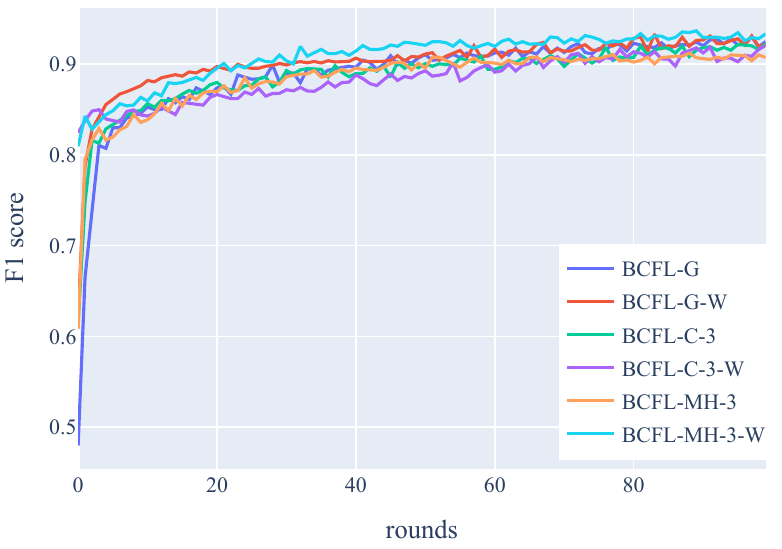}
        \caption{}
    \end{subfigure}
    \caption{Fashion-MNIST: (a) Accuracy (b) F-1 score (c) Accuracy warm-up comparison (d) F-1 score warm-up comparison.}
    \label{fig:fashion_acc_f1}
\end{figure}

\begin{figure*}[htb]
    \centering
    \begin{subfigure}[b]{0.26\textwidth}
        \includegraphics[width=\textwidth]{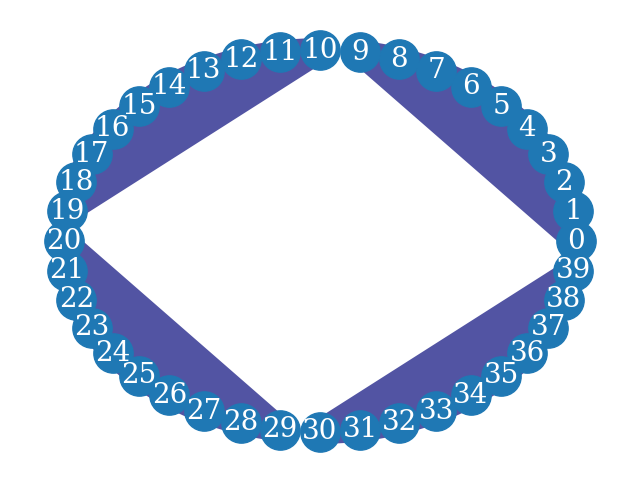}
        \caption{WeCFL}
    \end{subfigure}
    \hfill
    \begin{subfigure}[b]{0.26\textwidth}
        \includegraphics[width=\textwidth]{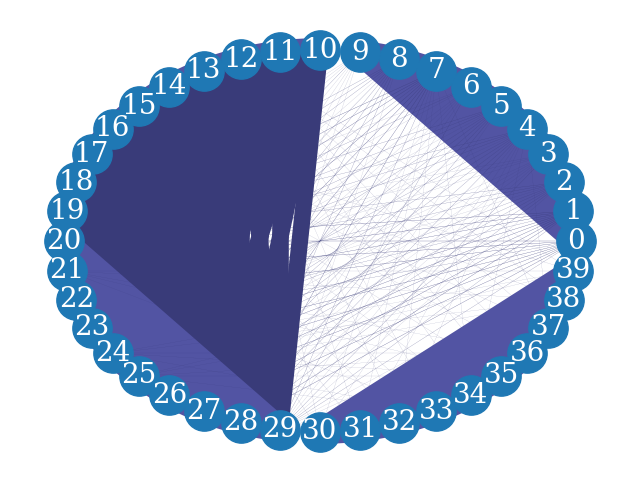}
        \caption{BCFL-G}
    \end{subfigure}
    \hfill
    \begin{subfigure}[b]{0.26\textwidth}
        \includegraphics[width=\textwidth]{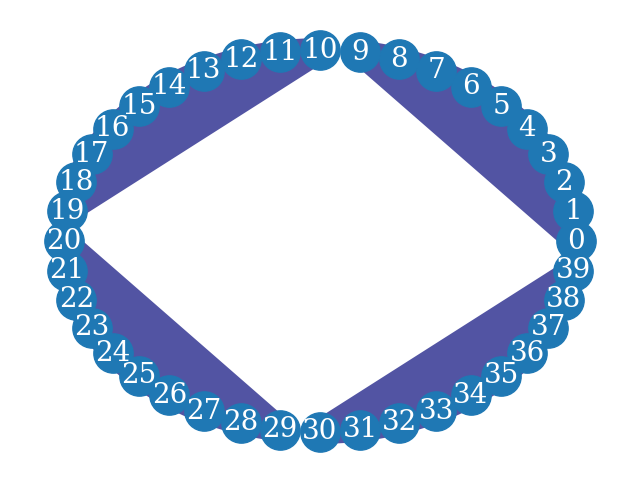}
        \caption{BCFL-G-W}
    \end{subfigure}
    \hfill
    \begin{subfigure}[b]{0.26\textwidth}
        \includegraphics[width=\textwidth]{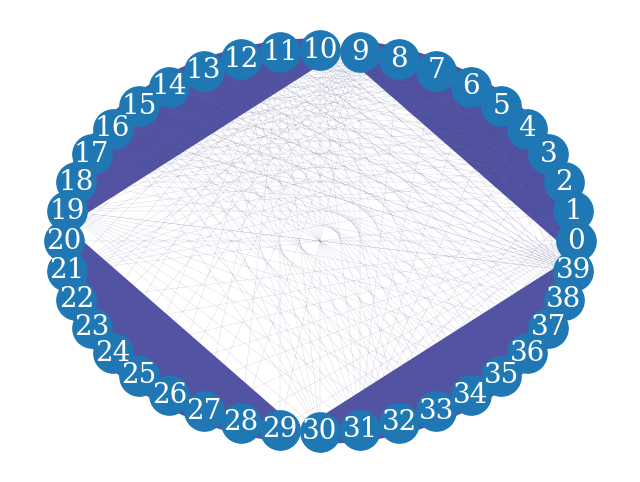}
        \caption{BCFL-C-3}
    \end{subfigure}
    \hfill
    \begin{subfigure}[b]{0.26\textwidth}
        \includegraphics[width=\textwidth]{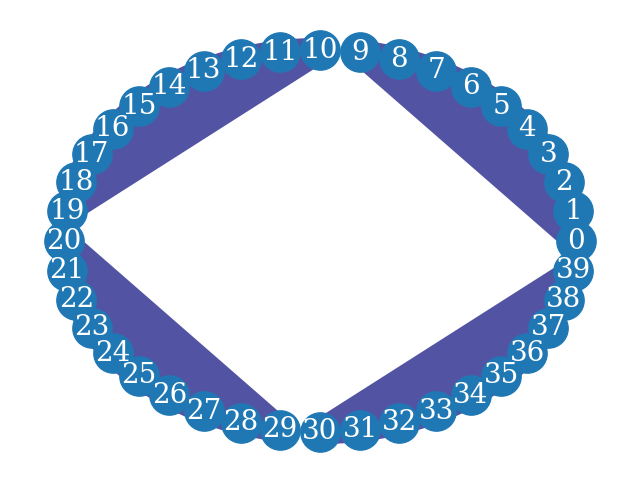}
        \caption{BCFL-C-3-W}
    \end{subfigure}
    \hfill
    \begin{subfigure}[b]{0.26\textwidth}
        \includegraphics[width=\textwidth]{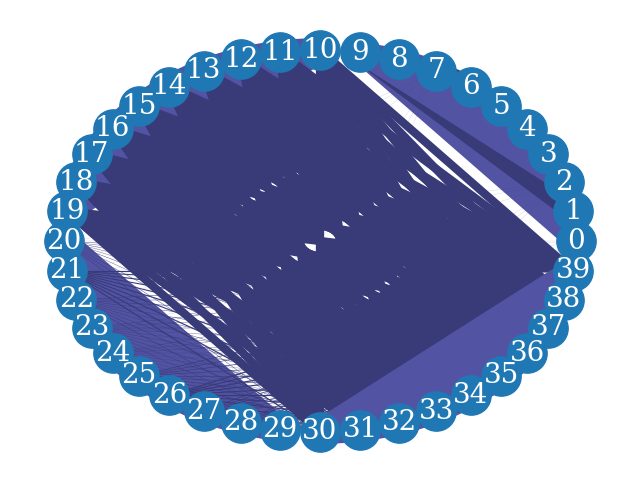}
        \caption{BCFL-C-6}
    \end{subfigure}
    \hfill
    \begin{subfigure}[b]{0.26\textwidth}
        \includegraphics[width=\textwidth]{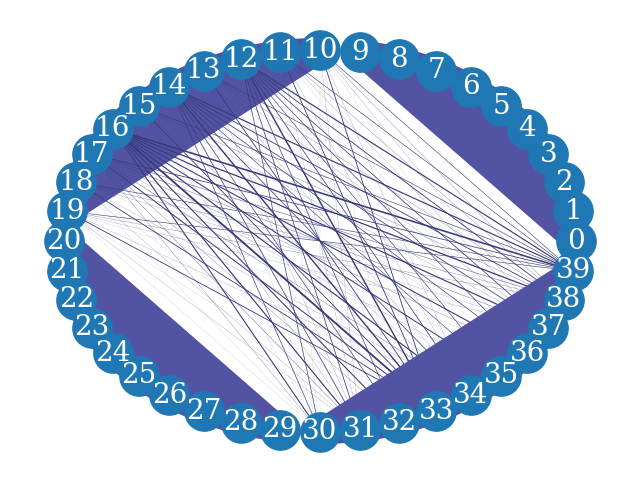}
        \caption{BCFL-C-6-W}
    \end{subfigure}
    \hfill
    \begin{subfigure}[b]{0.26\textwidth}
        \includegraphics[width=\textwidth]{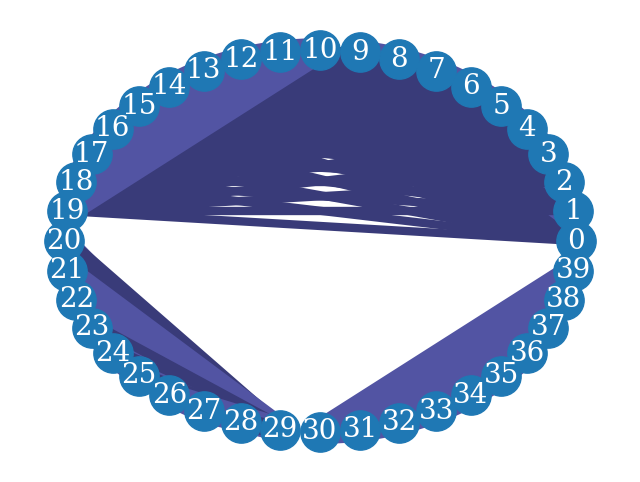}
        \caption{BCFL-MH-3}
    \end{subfigure}
    \hfill
    \begin{subfigure}[b]{0.26\textwidth}
        \includegraphics[width=\textwidth]{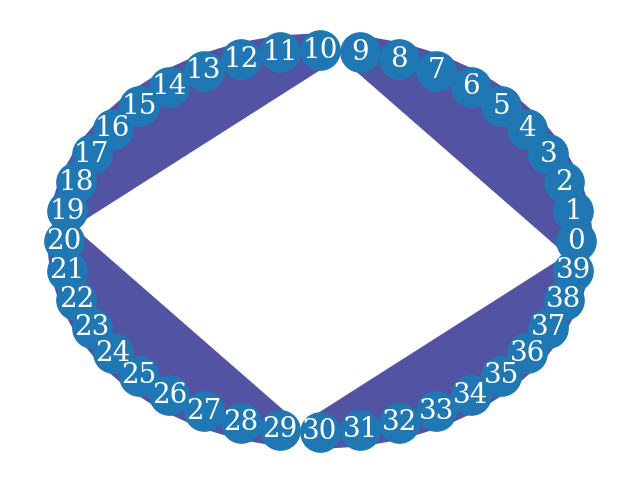}
        \caption{BCFL-MH-3-W}
    \end{subfigure}
    \hfill
    \begin{subfigure}[b]{0.26\textwidth}
        \includegraphics[width=\textwidth]{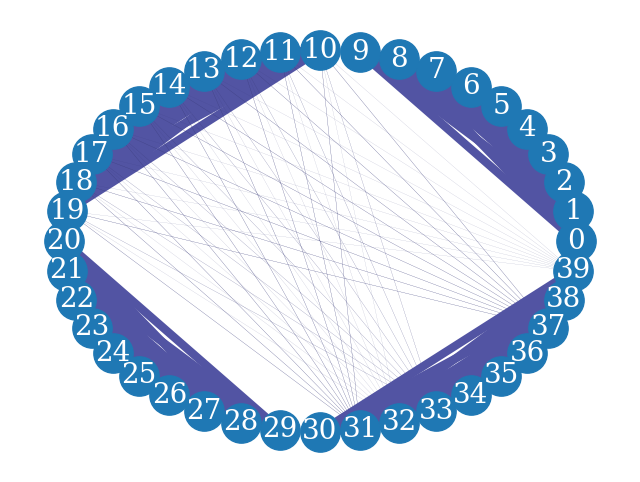}
        \caption{BCFL-MH-6}
    \end{subfigure}
    \hfill
    \begin{subfigure}[b]{0.26\textwidth}
        \includegraphics[width=\textwidth]{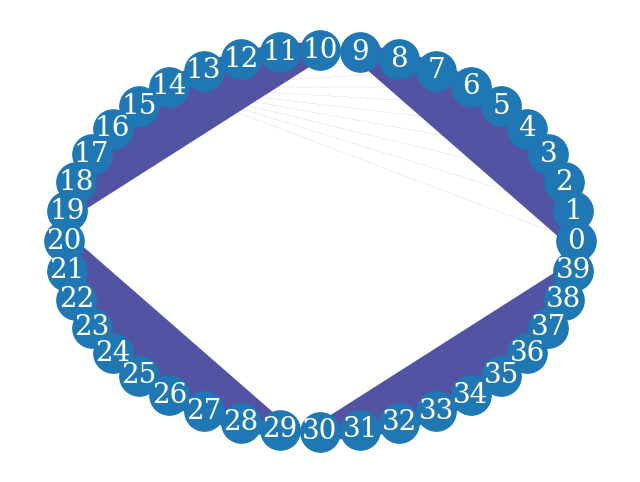}
        \caption{BCFL-MH-6-W}
    \end{subfigure}
    \caption{Client clustering during training for Fashion-MNIST dataset. Most of the time, clients are clustered in $4$ groups, which is consistent with the experiment setup of $K=4$ data groups with $C/K=10$ clients per group.}
    \label{fig:cluster_fashion}
\end{figure*}

\begin{figure}
    \centering
    \begin{subfigure}[b]{0.24\textwidth}
        \includegraphics[width=\textwidth]{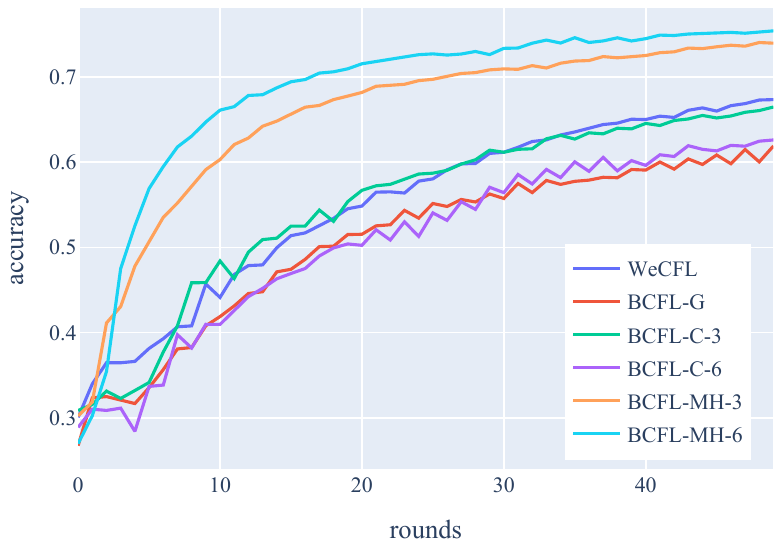}
        \caption{}
    \end{subfigure}
    \begin{subfigure}[b]{0.24\textwidth}
        \includegraphics[width=\textwidth]{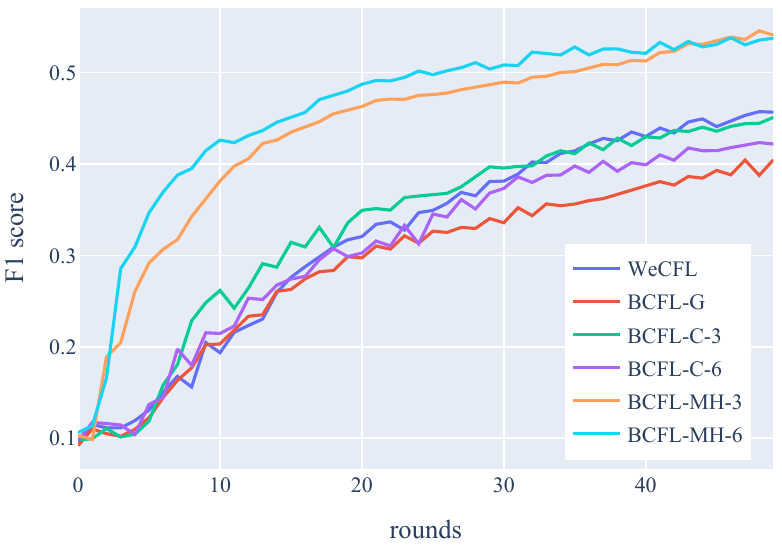}
        \caption{}
    \end{subfigure}
    \begin{subfigure}[b]{0.24\textwidth}
        \includegraphics[width=\textwidth]{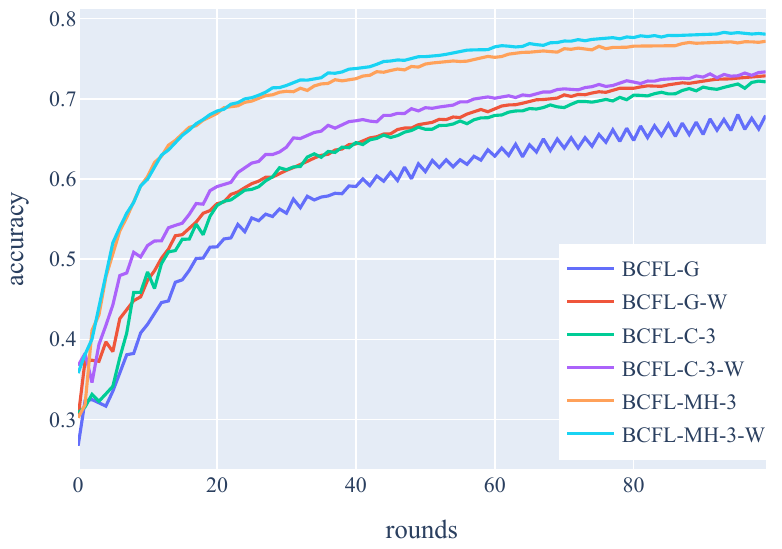}
        \caption{}
    \end{subfigure}
    \begin{subfigure}[b]{0.24\textwidth}
        \includegraphics[width=\textwidth]{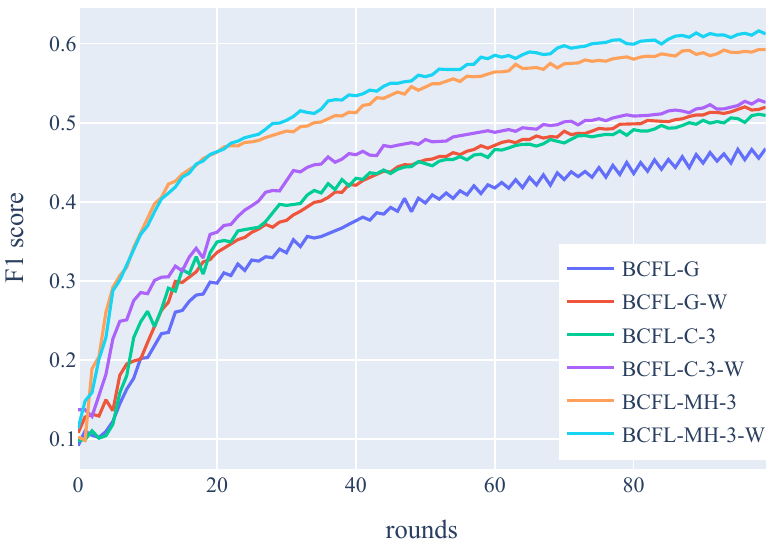}
        \caption{}
    \end{subfigure}
    \caption{CIFAR-10: (a) Accuracy (b) F-1 score (c) Accuracy warm-up comparison (d) F-1 score warm-up comparison.}
    \label{fig:cifar_acc_f1}
\end{figure}

\begin{figure*}[htb]
    \centering
    \begin{subfigure}[b]{0.26\textwidth}
        \includegraphics[width=\textwidth]{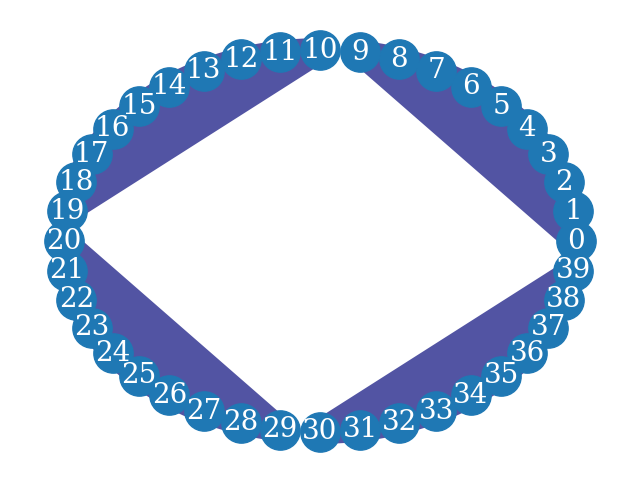}
        \caption{WeCFL}
    \end{subfigure}
    \hfill
    \begin{subfigure}[b]{0.26\textwidth}
        \includegraphics[width=\textwidth]{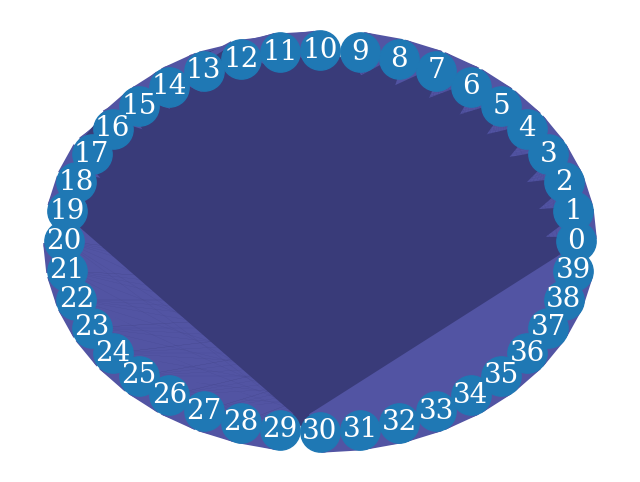}
        \caption{BCFL-G}
    \end{subfigure}
    \hfill
    \begin{subfigure}[b]{0.26\textwidth}
        \includegraphics[width=\textwidth]{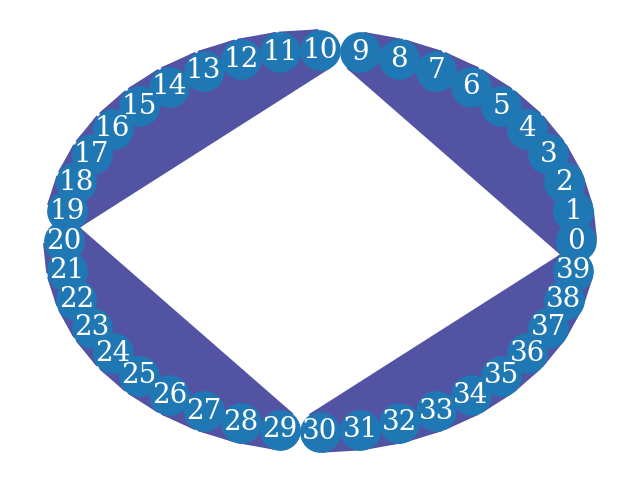}
        \caption{BCFL-G-W}
    \end{subfigure}
    \hfill
    \begin{subfigure}[b]{0.26\textwidth}
        \includegraphics[width=\textwidth]{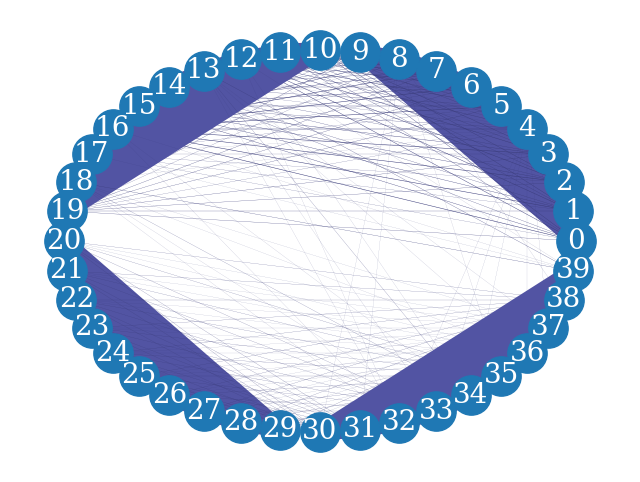}
        \caption{BCFL-C-3}
    \end{subfigure}
    \hfill
    \begin{subfigure}[b]{0.26\textwidth}
        \includegraphics[width=\textwidth]{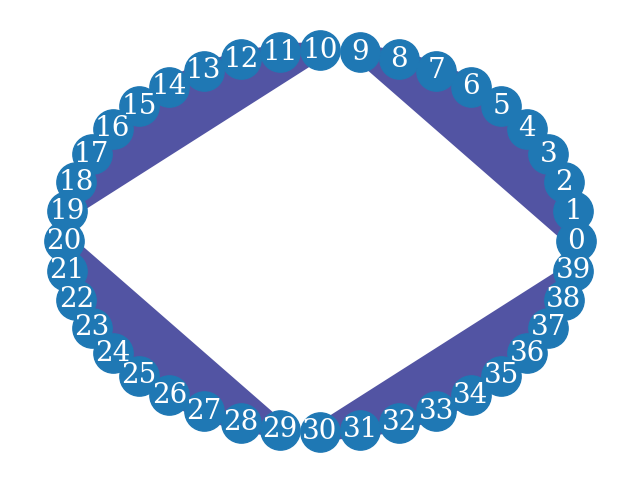}
        \caption{BCFL-C-3-W}
    \end{subfigure}
    \hfill
    \begin{subfigure}[b]{0.26\textwidth}
        \includegraphics[width=\textwidth]{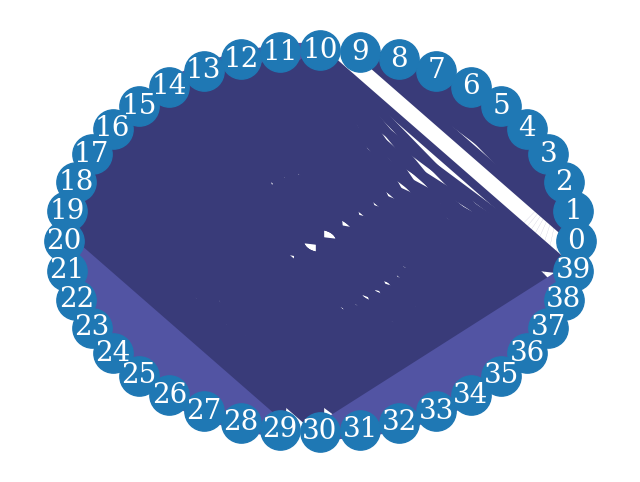}
        \caption{BCFL-C-6}
    \end{subfigure}
    \hfill
    \begin{subfigure}[b]{0.26\textwidth}
        \includegraphics[width=\textwidth]{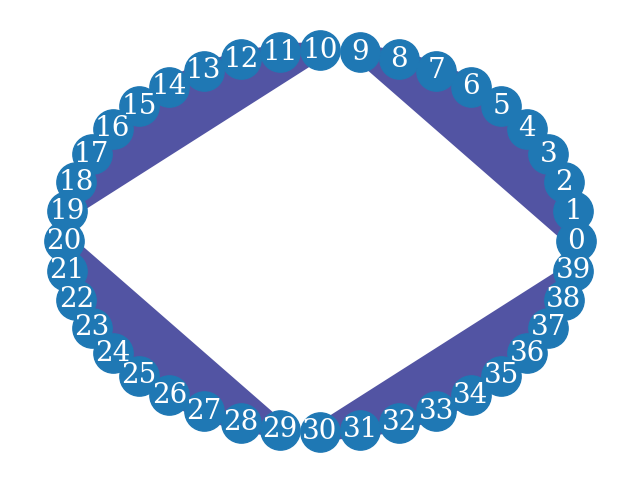}
        \caption{BCFL-C-6-W}
    \end{subfigure}
    \hfill
    \begin{subfigure}[b]{0.26\textwidth}
        \includegraphics[width=\textwidth]{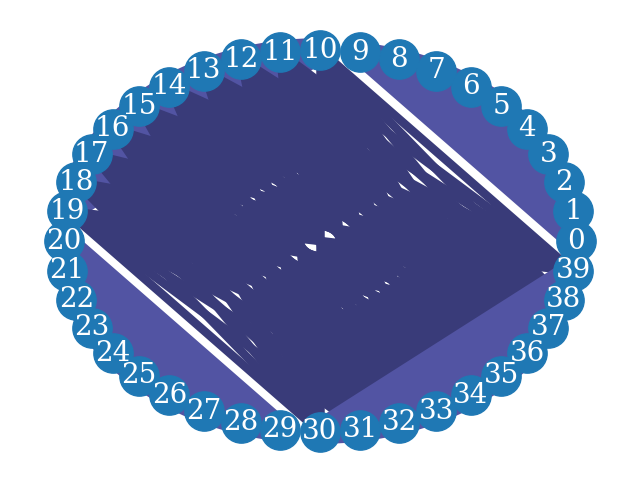}
        \caption{BCFL-MH-3}
    \end{subfigure}
    \hfill
    \begin{subfigure}[b]{0.26\textwidth}
        \includegraphics[width=\textwidth]{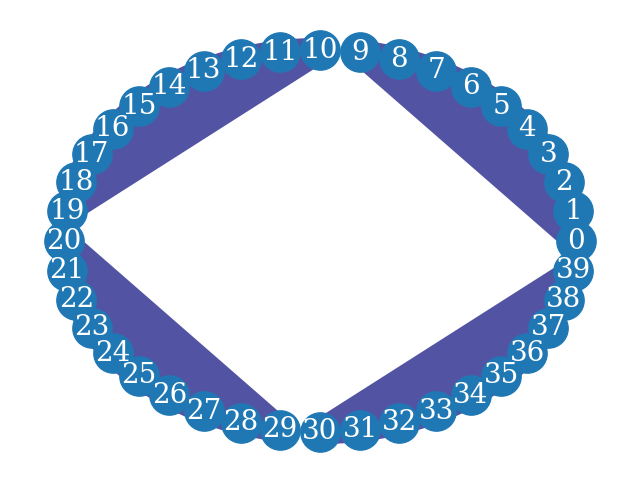}
        \caption{BCFL-MH-3-W}
    \end{subfigure}
    \hfill
    \begin{subfigure}[b]{0.26\textwidth}
        \includegraphics[width=\textwidth]{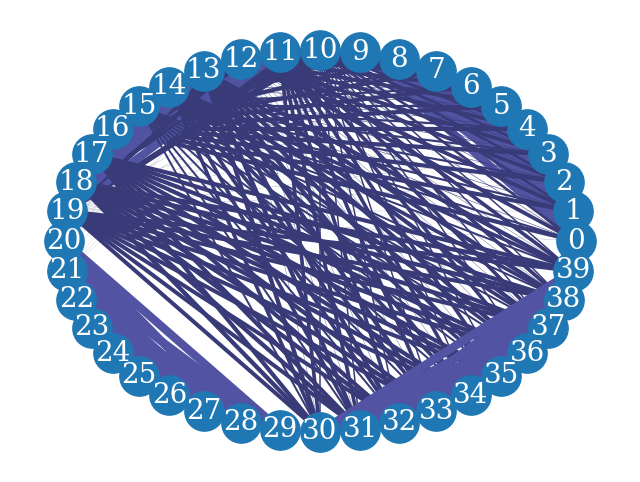}
        \caption{BCFL-MH-6}
    \end{subfigure}
    \hfill
    \begin{subfigure}[b]{0.26\textwidth}
        \includegraphics[width=\textwidth]{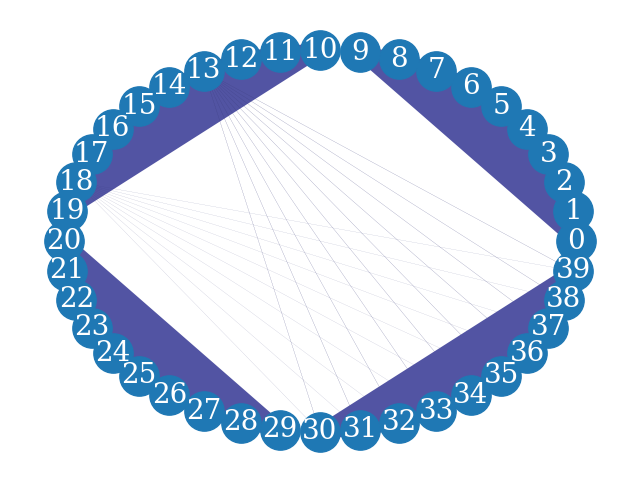}
        \caption{BCFL-MH-6-W}
    \end{subfigure}
    \caption{Client clustering during training for CIFAR-10 dataset. Most of the time, clients are clustered in $4$ groups, consistent with the experiment setup of $K=4$ data groups with $C/K=10$ clients per group.}
    \label{fig:cluster_cifar}
\end{figure*}

In addition to Tables 
, we provide additional results in this appendix. Notice that the analysis of all the results across the four experiments yields similar conclusions as discussed in Section \ref{sec:results} in terms of the superiority of BCFL, the ranking of the approximate solutions, as well as the ability to handle clustering hypotheses and associated uncertainty. This is a brief summary of those results:
\begin{itemize}
    \item Figure \ref{fig:digit5_acc_f1} shows the Digits-Five accuracy and F1-score results. Comparing both with/without warm-up initializations, we can see that the effect of incliding it is not dramatic. 
    \item Figure \ref{fig:cluster_digit5} shows the Digits-Five clustering results. We can observe very consistent clustering of clients observing similar data, as well as cross-pollination from other clients in BCFL schemes which helps improve the model training.
    \item Figure \ref{fig:amazon_acc_f1} shows the AmazonReview accuracy and F1-score results.
    \item Figure \ref{fig:cluster_amazon} shows the AmazonReview clustering results. AmazonReview data is text data containing user's reviews from a variety of products. Since the data distribution across clients is not extremely different, we can observe that even WeCFL connects more clients and therefore shares more information in the training phase. In this case, the method cannot easily cluster the data into four groups. In the case of BCFL, although it becomes more challenging, we can observe certain patterns and stronger associations among clients that should be clustered together.
    \item Figure \ref{fig:fashion_acc_f1} shows the Fashion-MNIST accuracy and F1-score results.
    \item Figure \ref{fig:cluster_fashion} shows the clustering results for Fashion-MNIST. Most of the time, the clients are correctly grouped into $4$ clusters, but they can also be grouped differently, especially with the proposed BCFL methods. This flexibility allows them to share more information with each other. Warm-up models are less shareable compared to models without warm-up, as a good initialization can lead to faster local convergence.
    \item Figure \ref{fig:cifar_acc_f1} shows the CIFAR-10 accuracy and F1-score results.
    \item Figure \ref{fig:cluster_cifar} shows the CIFAR-10 clustering results, yielding similar conclusions as with the Fashion-MNIST dataset.
\end{itemize}



\end{document}